\documentclass[10pt,twocolumn,letterpaper]{article}

\usepackage{times}
\usepackage{xcolor}
\usepackage{graphicx}
\usepackage{amsmath,amssymb}
\usepackage{booktabs}
\usepackage[numbers,sort&compress]{natbib}
\usepackage[hyphens]{url}
\usepackage[format=plain,labelformat=simple,labelsep=period,font=small]{caption}
\usepackage{float,placeins}
\usepackage{cuted} 
\usepackage{tcolorbox}
\tcbuselibrary{skins}
\usepackage[shortlabels,inline]{enumitem}
\setlist[itemize,enumerate]{noitemsep,leftmargin=*,topsep=0em}

\renewenvironment{abstract}{%
  \centerline{\large\bfseries Abstract}%
  \vspace*{12pt}\noindent\itshape\ignorespaces
}{\par\vspace*{12pt}}

\makeatletter
\renewcommand{\@seccntformat}[1]{\csname the#1\endcsname.\enspace}
\renewcommand{\section}{\@startsection{section}{1}{\z@}%
  {-10pt plus -2pt minus -2pt}{7pt}{\normalfont\large\bfseries}}
\renewcommand{\subsection}{\@startsection{subsection}{2}{\z@}%
  {-8pt plus -2pt minus -2pt}{5pt}{\normalfont\fontsize{11}{13}\selectfont\bfseries}}
\renewcommand{\subsubsection}{\@startsection{subsubsection}{3}{\z@}%
  {-6pt plus -2pt minus -2pt}{3pt}{\normalfont\normalsize\bfseries}}

\let\await@body@addtocurcol\@addtocurcol
\let\await@body@combinedblfloats\@combinedblfloats
\usepackage{stfloats}
\let\await@appendix@addtocurcol\@addtocurcol
\let\await@appendix@combinedblfloats\@combinedblfloats
\let\@addtocurcol\await@body@addtocurcol
\let\@combinedblfloats\await@body@combinedblfloats
\newcommand{\EnableAppendixBottomFloats}{%
  \let\@addtocurcol\await@appendix@addtocurcol
  \let\@combinedblfloats\await@appendix@combinedblfloats
}
\makeatother

\definecolor{linkblue}{rgb}{0.21,0.49,0.74}
\usepackage[pagebackref,breaklinks,colorlinks,allcolors=linkblue,pdfusetitle]{hyperref}

\title{Action Windows Are Inherently Temporal: A Benchmark for Time-Sensitive Robotic Manipulation}

\newcommand{\paperauthors}{Haoxiang Zhan, Haoqin Hong, Congcong Zhu, Jingrun Chen}
\author{\paperauthors\\
  University of Science and Technology of China}
\hypersetup{pdfauthor={\paperauthors}}
\date{}

\begin{document}

\maketitle
\setlength{\stripsep}{10pt plus 2pt minus 2pt}
\begin{strip}
\centering
\includegraphics[width=0.93\textwidth]{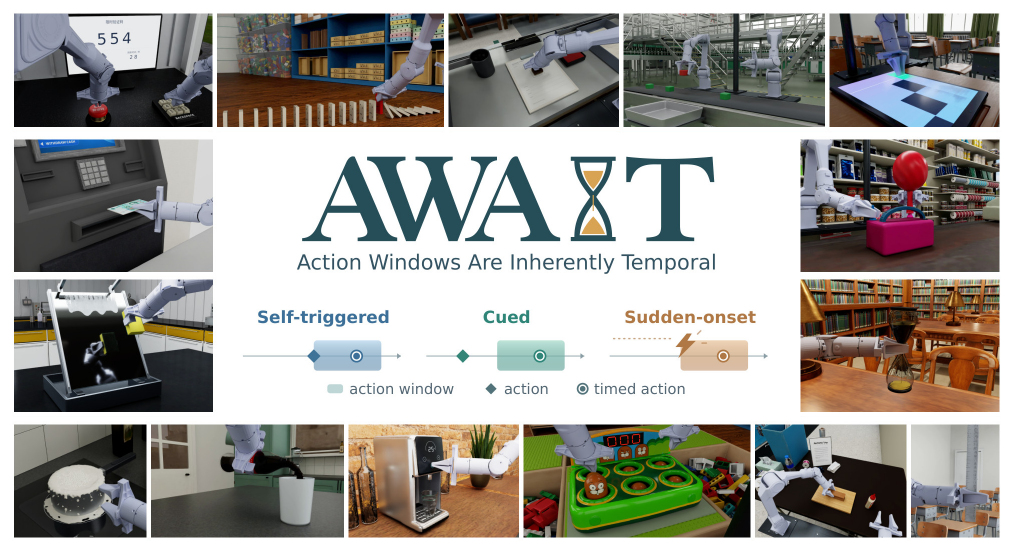}
\captionsetup{hypcap=false}
\captionof{figure}{\textbf{Overview of AWAIT.}
AWAIT evaluates robotic manipulation tasks whose goals must be achieved within action windows set by evolving processes.
The benchmark comprises 18 tasks spanning rigid-body interactions and complex physical processes, including liquid flow, inflation, and surface-state changes, in diverse simulated environments.
Our scene generation pipeline embeds these tasks in photorealistic 3D environments with varied object layouts.
The tasks fall into three window types based on the information available to agents about window onset: self-triggered, cued, and sudden-onset.}
\label{fig:teaser}
\captionsetup{hypcap=true}
\end{strip}

\begin{abstract}
Vision-language model (VLM) agents have surpassed learned-policy baselines on existing robotic manipulation benchmarks.
However, these agents still struggle with time-sensitive everyday tasks that existing benchmarks for VLM agents cover only partially.
In these tasks, evolving physical processes create finite windows for successful intervention. These action windows are inherently temporal and impose a shared time budget on decision making and physical execution.
Coordinating intervention within this budget requires agents to determine when the window opens.
We introduce \textbf{AWAIT}, a simulation benchmark that evaluates task completion within action windows while accounting for decision and execution time.
Our evaluation reveals difficulties in recognizing window onset and shows that successive decisions and movements can exhaust an action window even when each decision takes less than the window duration.
Project page: \url{https://eyja42.github.io/await-site/}.
\end{abstract}

\section{Introduction}
\label{sec:intro}

Everyday tasks often require intervening while physical conditions still permit success, whether bonding a handle before glue dries, stopping heating before milk overflows, or catching a falling cup before impact. Recent vision-language model (VLM) agents have expanded the capabilities available for robotic manipulation by combining general-purpose models with perception tools and executable programs~\cite{pyrua,q0preview}. As these agents achieve leading scores on a public manipulation benchmark~\cite{robodojo}, evaluating their capabilities in time-sensitive everyday tasks becomes increasingly important. Such evaluation must determine whether agents complete the required intervention in time to achieve the task goal.

\begin{figure*}[!t]
    \centering
    \includegraphics[width=\textwidth]{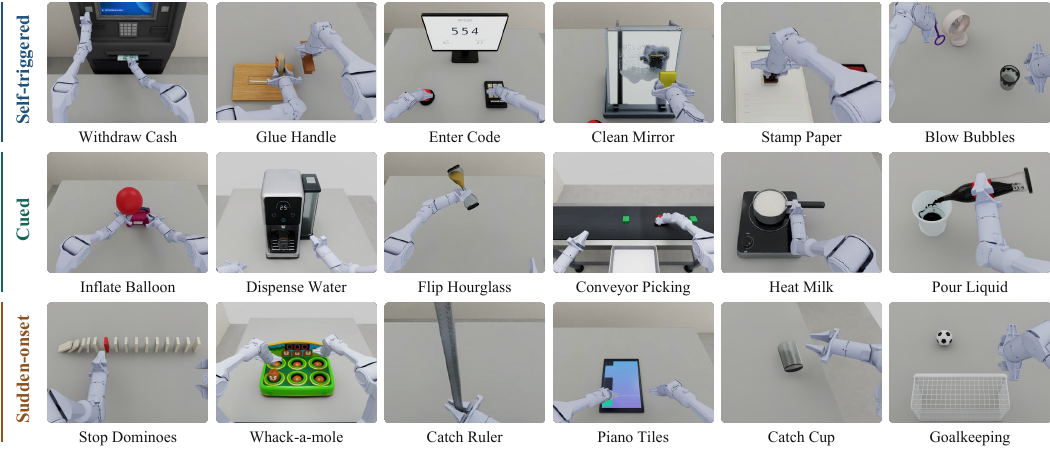}
    \caption{\textbf{AWAIT tasks, grouped by window-opening information.}
    AWAIT evaluates task completion within action windows, accounting for both computation and physical execution time.
    We use plain backgrounds to facilitate comparison of task objects and interactions.
    \textbf{Self-triggered (top):} a designated agent action opens the window, allowing preparation before triggering.
    \textbf{Cued (middle):} observable process changes provide advance cues for anticipating window onset.
    \textbf{Sudden-onset (bottom):} windows open without reliable advance cues, requiring event detection and timely response.}
    \label{fig:suite}
\end{figure*}

Recent benchmarks evaluate dynamic manipulation with ongoing environmental evolution and explicit inference delays~\cite{robodyna,reflexbench,motionforge}. Industrial manipulation evaluations also incorporate finite execution windows into task specifications~\cite{liberoindustry}. Understanding timely intervention further requires relating an agent's complete decision and execution sequence to the interval in which intervention can succeed. Task scores and individual decision times alone do not explain how agents use this interval.

For this analysis, we use \emph{action windows}: process-dependent intervals during which the required intervention can achieve the task goal (Fig.~\ref{fig:teaser}). Timely intervention requires recognizing window onset and completing the required operations before closure. Window duration provides the corresponding time budget for model inference, tool computation, and physical execution. Even when each decision takes less time than the window duration, successive decisions and movements can exhaust this budget.

We introduce \textbf{AWAIT}, a simulation benchmark comprising 18 tasks across self-triggered, cued, and sudden-onset windows (Fig.~\ref{fig:suite}). The tasks span rigid-body interactions, liquid transfer, inflation, and surface-state changes in photorealistic environments built with our scene generation pipeline. The evaluation protocol accounts for computation and physical execution on a shared simulation clock, with measured-time and controlled-delay conditions that preserve the same physical execution rules. We evaluate VLM agents across decision costs normalized by window duration and use execution traces to examine how their decisions and movements consume the available time.

Our contributions are:
\begin{itemize}
    \item \textbf{An action-window task suite} covering diverse physical processes with explicit window boundaries and task-specific success conditions.
    \item \textbf{A computation-aware evaluation protocol} that measures decision costs relative to available intervention time while preserving physical execution rules.
    \item \textbf{A behavioral analysis of timely intervention} linking decision and motion sequences to window use and task outcomes.
\end{itemize}

\section{Related Work}
\label{sec:related}

\paragraph{Benchmarks with physical interactions. }
Manipulation benchmarks represent physical interactions through material dynamics and object-state changes. SoftGym provides tasks for deformable-object manipulation~\cite{softgym}, and FluidLab benchmarks interactions between solids and fluids and among multiple fluids~\cite{fluidlab}. BEHAVIOR-1K incorporates object-state changes into household activities~\cite{behavior1k}, and ARNOLD specifies continuous goal states for language-grounded manipulation~\cite{arnold}.

\paragraph{Time-sensitive benchmarks.}
Dynamic manipulation benchmarks examine object motion, process constraints, and sensitivity to decision delays. DOMINO and the DOM benchmark introduced with DynamicVLA evaluate manipulation of moving objects~\cite{domino,dynamicvla}. LIBERO-Industry varies conveyor speed, interaction-window duration, and geometric tolerances~\cite{liberoindustry}. RoboDyna compares synchronous and asynchronous protocols under measured inference latency~\cite{robodyna}, ReflexBench supports prescribed and measured inference delays~\cite{reflexbench}, and MotionForge advances the environment independently of policy inference during long-horizon evaluation~\cite{motionforge}. LIBERO-Dynamic jointly examines target motion, inference latency, and action-execution strategies~\cite{liberodynamic}. Related temporal-reasoning benchmarks address concurrent, durative actions in Robotouille~\cite{robotouille}, interval timing in Chronocooked~\cite{chronocooked}, and memory-dependent manipulation in LIBERO-Memory~\cite{dysta}.

\paragraph{Language-model agents for manipulation.}
Language-model agents organize execution through skill selection and program generation. SayCan selects predefined skills using learned affordance value functions~\cite{saycan}; Code as Policies makes the execution logic programmable~\cite{cap}, and VoxPoser uses generated code to compose visual value maps for closed-loop trajectory planning~\cite{voxposer}. Generated programs also support feedback during execution: PyRUA-Lean checks and retries through Python cells~\cite{pyrua}, while Q0-Preview combines concurrent planning with code tools and execution checks~\cite{q0preview}. Agentic GR00T~2 combines a world-action model and code-based skills under a reasoning agent~\cite{gr00t2}. CaP-X compares coding agents for manipulation~\cite{capx}, while RobotUse studies the organization of computation, context, and decisions during robot execution~\cite{robotuse}.

\paragraph{Environment construction for manipulation.}
Environment construction combines scene geometry with the interactions needed for manipulation. Gaussian-splatting methods reconstruct observed workspaces for simulation~\cite{splatsim,robogsim}, while asset-generation methods produce objects with physical or articulated structure~\cite{physxanything,infinigenarticulated}. HomeWorld assembles furnished, interactive environments from floorplans~\cite{homeworld}, and GRS generates task simulations and test scripts from scene images~\cite{grs}.

\smallskip
\noindent\textbf{AWAIT} complements benchmarks of dynamic manipulation and inference latency by relating complete decision and execution sequences to process-defined action windows. We group tasks by window-opening information to examine how agents decide when to intervene. Trajectory analysis tracks the time consumed by successive decisions and movements within these windows.

\section{AWAIT Task Suite and Evaluation Protocol}
\label{sec:suite}

\subsection{Tasks and action windows}
\label{sec:suite:windows}

An episode can contain one or more \emph{events}, each associated with a task-defined goal and an action window. Achieving the goal may require an ordered sequence of operations or sustained interaction with the process. Preparation can begin before the window opens; an event is judged successful only if its goal is achieved within the window.

Let $s_t$ denote the world state at simulation time $t$. For an event $e$, the binary predicate $\varphi^e_{\mathrm{open}}(s_t)$ tests the task-defined opening condition. When this condition is reached, the opening time is
\begin{equation}
t_{\mathrm{open}}(e)=\inf\{t:\varphi^e_{\mathrm{open}}(s_t)=1\}.
\label{eq:topen}
\end{equation}

The closing time is defined through a reference continuation without further intervention by the evaluated system. For a given instance, this continuation starts from the actual opening state and follows the task's prescribed process dynamics. Denote the reference world state by $\bar{s}_t^{\,e}$, with $\bar{s}_{t_{\mathrm{open}}(e)}^{\,e}=s_{t_{\mathrm{open}}(e)}$. The binary predicate $\varphi^e_{\mathrm{close}}$ indicates that the process has left the task-defined conditions for valid goal completion. The closing time is
\begin{equation}
t_{\mathrm{close}}(e)=\inf\{t\geq t_{\mathrm{open}}(e):\varphi^e_{\mathrm{close}}(\bar{s}_t^{\,e})=1\}.
\label{eq:tclose}
\end{equation}
The action window is $W(e)=[t_{\mathrm{open}}(e),t_{\mathrm{close}}(e))$, with duration $|W(e)|=t_{\mathrm{close}}(e)-t_{\mathrm{open}}(e)$. Once opening occurs, the reference continuation fixes the closing time independently of subsequent system actions. Successful interaction can stop the actual process while the complete window remains defined.

Let $t_{\mathrm{goal}}(e)$ denote when the event's goal is achieved along the actual trajectory, satisfying the task's state conditions and required history. Event success requires $t_{\mathrm{goal}}(e)\in W(e)$.

\subsection{Window types and task coverage}
\label{sec:suite:tasks}

\begin{figure}[!ht]
\centering
\includegraphics[width=\columnwidth]{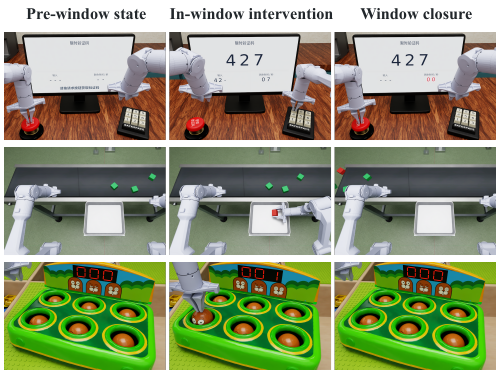}
\caption{\textbf{Consequences of missing an action window.}
Illustrative simulator views show \emph{Enter Code}, \emph{Conveyor Picking}, and \emph{Whack-a-mole} (top to bottom).
The columns show the pre-window state, intervention within the window, and window closure.
The right column illustrates what would happen without timely intervention: the code expires, the target leaves the pickup region, or the mole retracts, closing the opportunity for successful completion of the corresponding event.}
\label{fig:action-windows}
\end{figure}

The suite contains 18 tasks, with six per window type (Fig.~\ref{fig:suite}). Each task is assigned to one type according to how the system obtains information about when $\varphi^e_{\mathrm{open}}(s_t)$ becomes true. The associated window duration follows from the task's process dynamics and opening state. Figure~\ref{fig:action-windows} illustrates timely intervention and the consequences of missing an action window in three representative tasks.

\paragraph{Self-triggered windows.}
A designated trigger action by the evaluated system opens the target window. The tasks are \emph{Withdraw Cash}, \emph{Glue Handle}, \emph{Enter Code}, \emph{Clean Mirror}, \emph{Stamp Paper}, and \emph{Blow Bubbles}. Preparation before triggering and timely completion of the subsequent operations contribute to success.

\paragraph{Cued windows.}
The target window opens as a process evolves, with observable cues available before opening. The tasks are \emph{Inflate Balloon}, \emph{Dispense Water}, \emph{Flip Hourglass}, \emph{Conveyor Picking}, \emph{Heat Milk}, and \emph{Pour Liquid}. Cues such as liquid level and balloon size can guide the timing of the required operations; initiating the process does not itself open the target window.

\paragraph{Sudden-onset windows.}
The target window opens without a reliable observable precursor. The tasks are \emph{Stop Dominoes}, \emph{Whack-a-mole}, \emph{Catch Ruler}, \emph{Piano Tiles}, \emph{Catch Cup}, and \emph{Goalkeeping}. After onset, observations of the evolving process may still guide task execution.

\subsection{Scene generation and task validation}
\label{sec:scene}

Given reference images or layout descriptions, our scene generation pipeline constructs photorealistic 3D environments and integrates task assets, mechanisms, and the robot and camera rig. TRELLIS.2~\cite{trellis2} converts generated asset reference images into textured meshes for background furnishings and task objects. Task objects are modeled and textured in greater detail than background furnishings and are manually inspected by the authors. Blender defines precise dimensions, openings, and mating surfaces, with optional UltraShape~\cite{ultrashape} refinement. Backgrounds can be replaced while retaining task objects, equipment, and mechanisms. GPT-6 Astra places assets with robot access and camera visibility as constraints. Each instance's initial layout and task parameters remain fixed across evaluations. Domain randomization keeps parameters governing action-window duration fixed. Within these constraints, realized durations can still vary with the opening state.

\begin{figure}[!htbp]
\centering
\includegraphics[width=\columnwidth]{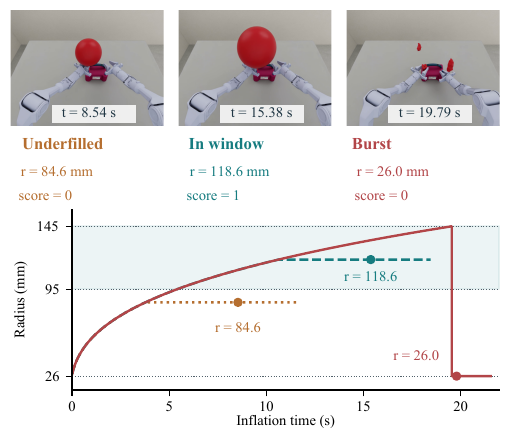}
\caption{\textbf{A process-defined action window in Inflate Balloon.}
The window opens at the target radius and closes when continued inflation would cause bursting; the corresponding radius range is shaded.
The three examples show underfilling, successful completion within the window, and bursting.}
\label{fig:state}
\end{figure}

Assets are exported to USD with collision geometry, rigid-body properties, and joints where required. Task mechanisms extend these assets with evolving material states. We define particle-based liquids through PhysX USD schemas and simulate them using the PhysX position-based dynamics (PBD) solver in Isaac Sim. Liquid transfer and absorption are measured from the evolving particle state. Task-specific process models describe temperature-driven liquid expansion and foam formation, gas-volume changes during inflation, and surface deposition, drying, and bonding. These processes respond to measured contacts and object poses, allowing robot interaction to alter their evolution. The same process states determine visible changes in liquid level, foam, geometry, or surface appearance and supply the predicates used to define action windows and evaluate intervention (Fig.~\ref{fig:state}).

After a settling check confirms stable support and no collisions, each assembled instance undergoes task validation using a robot trajectory constructed from task specifications and privileged information about object positions. Physics and task mechanisms remain active throughout validation. Acceptance requires the prescribed goal and timing conditions to be satisfied and the judge to award the full episode score before the episode horizon. Failed configurations are corrected and revalidated. The supplementary material provides construction details and further examples of generated environments.

\subsection{Interaction and time accounting}
\label{sec:suite:clock}

We evaluate each method as a black-box system encompassing its model, perception and planning tools, and execution routines. Through a common interface, the system receives public task instructions, scene observations, and robot feedback and submits control targets. Under the standard protocol, an access boundary isolates every system component from host-side task code, judges, and \emph{privileged information} obtained by directly querying world state, such as object positions. The host uses privileged information for scene construction, instance validation, and judging; diagnostic experiments that expose such information to a method are configured and reported separately.

Robot motion and scene dynamics advance on the simulation clock, whereas method computation is timed in real time. The protocol translates computation into simulation-time costs using either measured-time or controlled-delay accounting. In both conditions, physical execution advances simulation by the requested control duration, with the same physics step and actuator limits.

A high-level decision round determines the next motion command or control program and may include multiple model calls and tool computations. The resulting program can use feedback to update control targets without starting a new high-level round for each update. For round $k$, the computation cost in simulation seconds is
\begin{equation}
d_k=\begin{cases}
\Delta t^{\mathrm{real}}_k/\rho & \text{measured-time condition},\\
\tau & \text{controlled-delay condition}.
\end{cases}
\label{eq:decision-cost}
\end{equation}
Here, $\Delta t^{\mathrm{real}}_k$ is the adjusted elapsed real time for round $k$, covering model calls, perception, planning, and program computation. The adjustment excludes designated API response-wait intervals while retaining overlapping computation. In the measured-time condition, $\rho>0$ is the computation speedup factor; computation outside high-level rounds, including feedback processing during program execution, follows the same adjustment and $1/\rho$ conversion. In the controlled-delay condition, each high-level round incurs a fixed delay $\tau>0$ upon completion, regardless of real computation time, and feedback-driven updates within the resulting program incur no additional computation cost.

Before new control targets take effect, the simulation advances by the accounted computation time while task processes evolve and the robot tracks the previously active targets. Observation and target submission do not themselves advance simulation time, and command submission times are recorded separately from goal achievement times. Both computation costs and physical execution thus consume the available intervention time, and action windows can open or close during either. The supplementary material specifies how API response waits are deducted while accounting for concurrent computation.

\setcounter{dbltopnumber}{1}

\subsection{Scoring and reporting}
\label{sec:suite:metrics}

 A single-event task receives a score of 1 for success and 0 for failure. For tasks with multiple events, the episode score combines event outcomes and task-defined error penalties and is normalized to $[0,1]$.

To compare decision delays across tasks, we calibrate a reference-window length $W^\star$ from each task's fixed privileged validation trajectories (Sec.~\ref{sec:scene}). These trajectories supply event-opening states; the prescribed reference continuations determine the corresponding window lengths. We average lengths within each instance, then weight instances equally. The resulting task-level $W^\star$ is shared across methods, models, and timing conditions, independently of their rollout outcomes. The normalized delay $\tau/W^\star$ expresses a high-level decision cost relative to this calibration scale; individual event windows can differ with the actual opening state. The supplementary material details calibration and task-specific reference windows.

Averaging the task-level reference values equally within each type gives 21.8\,s for self-triggered tasks, 18.1\,s for cued tasks, and 2.2\,s for sudden-onset tasks. These averages summarize the time scales represented in the suite; delay normalization uses each task's own $W^\star$ to account for differences within a type.

\begin{table*}[t]

\centering
\caption{Measured-time comparison: mean task score (\%). In teleoperation, $\rho$ gives the seconds of real time per simulation second. For automated methods, $\rho$ scales the adjusted time defined in Sec.~\ref{sec:suite:clock}. $\dagger$ marks the default VLMs specified in the prior methods' open-source repositories. The best and second-best Overall scores at $\rho=10$ and $\rho=100$ are highlighted in bold. Scores are rounded to one decimal place after aggregation.}
\label{tab:main}
\small
\setlength{\tabcolsep}{2.2pt}
\renewcommand{\arraystretch}{1.10}
\begin{tabular*}{\textwidth}{@{\extracolsep{\fill}}ll*{12}{r}@{}}
\toprule
& & \multicolumn{4}{c}{$\rho=1$} & \multicolumn{4}{c}{$\rho=10$} & \multicolumn{4}{c}{$\rho=100$} \\
\cmidrule(lr){3-6}\cmidrule(lr){7-10}\cmidrule(lr){11-14}
Method & VLM & Self & Cued & Sudden & Overall & Self & Cued & Sudden & Overall & Self & Cued & Sudden & Overall \\
\midrule
Native agent & GPT-6 Astra / high & 0.0 & 0.0 & 0.0 & 0.0 & 30.6 & 36.1 & 3.7 & 23.5 & 30.6 & 55.6 & 47.1 & 44.4 \\
             & GPT-6.1 Sol / xhigh & 0.0 & 0.0 & 0.0 & 0.0 & 27.8 & 30.6 & 11.6 & 23.3 & 41.7 & 44.4 & 52.3 & \textbf{46.1} \\
\addlinespace[2pt]
CaP-X~\cite{capx} & Gemini 3.1 Pro / high$^\dagger$ & 0.0 & 0.0 & 0.0 & 0.0 & 0.0 & 0.0 & 0.0 & 0.0 & 8.3 & 2.8 & 1.7 & 4.3  \\
    & GPT-6.1 Sol / xhigh & 0.0 & 0.0 & 0.0 & 0.0 & 0.0 & 0.0 & 0.0 & 0.0 & 19.4 & 8.3 & 2.8 & 10.2  \\
\addlinespace[2pt]
PyRUA-Lean~\cite{pyrua} & GPT-6 Astra / high$^\dagger$ & 0.0 & 0.0 & 0.0 & 0.0 & 50.0 & 44.4 & 13.8 & \textbf{36.1} & 55.6 & 50.0 & 46.8 & \textbf{50.8} \\
           & GPT-6.1 Sol / xhigh & 0.0 & 0.0 & 0.0 & 0.0 & 30.6 & 41.7 & 6.0 & \textbf{26.1} & 38.9 & 50.0 & 36.1 & 41.7 \\
\addlinespace[2pt]
Q0-Preview~\cite{q0preview} & Claude Opus 5.5 / xhigh$^\dagger$ & 0.0 & 0.0 & 0.0 & 0.0 & 13.9 & 19.4 & 2.8 & 12.0 & 50.0 & 55.6 & 25.2 & 43.6 \\
           & GPT-6.1 Sol / xhigh & 0.0 & 0.0 & 0.0 & 0.0 & 16.7 & 22.2 & 0.5 & 13.1 & 58.3 & 52.8 & 23.6 & 44.9 \\
\addlinespace[2pt]
Human (arm teleop) & N/A & 91.1 & 97.8 & 67.2 & 85.4 & 100.0 & 100.0 & 79.9 & 93.3 & 100.0 & 100.0 & 100.0 & 100.0 \\
\bottomrule
\end{tabular*}

\end{table*}

\section{Experiments}
\label{sec:exp}
We evaluate VLM agents on AWAIT across decision costs, then examine how window information and the organization of decision and execution sequences affect timely task completion.

\subsection{Experimental setup}
\label{sec:exp:setup}

\paragraph{Evaluated methods.}
We evaluate four methods as black-box systems under the protocol in Sec.~\ref{sec:suite:clock}, retaining their respective planning and execution structures. Native agent uses the VLM's tool-calling capabilities to select arm and gripper targets directly from RGBD observations and measured robot state. The targets are executed through the public robot controller, without a code-execution environment or a task-specific planner. CaP-X~\cite{capx} generates control programs with multi-turn visual feedback, PyRUA-Lean~\cite{pyrua} interleaves Python cells with execution feedback, and Q0-Preview~\cite{q0preview} combines concurrent planning, code tools, and execution checks. We evaluate the three prior methods with the default VLMs specified in their respective open-source repositories and use GPT-6 Astra with high reasoning effort for Native agent. To compare responses to decision-time cost under a shared VLM, we also evaluate each method with GPT-6.1 Sol at xhigh reasoning effort.

\begin{table}[b]
\centering
\caption{\textbf{High-level decision round latency.} Mean adjusted time with GPT-6.1 Sol/xhigh. Means are rounded to two decimal places after aggregation.}
\label{tab:exp:latency}
\small
\setlength{\tabcolsep}{3pt}
\begin{tabular*}{\columnwidth}{@{\extracolsep{\fill}}lr@{}}
\toprule
Method & \shortstack[r]{Mean time per high-level\\decision round (s)} \\
\midrule
Native agent & 11.96 \\
CaP-X~\cite{capx} & 130.80 \\
PyRUA-Lean~\cite{pyrua} & 26.88 \\
Q0-Preview~\cite{q0preview} & 70.22 \\
\bottomrule
\end{tabular*}

\end{table}

\begin{table*}[t]
\centering
\caption{Mean task score (\%) by tertile of normalized delay $\tau/W^\star$. $\dagger$ marks the default VLMs specified in the prior methods' open-source repositories. Method--VLM configurations share the pooled tertile boundaries. Within each group, scores at different delays are first averaged for each task; type scores and Overall then average the corresponding tasks with equal weights. The best and second-best Overall scores are highlighted in bold. Scores are rounded to one decimal place after aggregation.}
\label{tab:exp:controlled}
\small
\setlength{\tabcolsep}{1.8pt}
\renewcommand{\arraystretch}{1.10}
\begin{tabular*}{\textwidth}{@{\extracolsep{\fill}}ll*{12}{r}@{}}
\toprule
& & \multicolumn{4}{c}{\shortstack{Lower tertile\\{\footnotesize($\tau/W^\star<6\%$)}}} & \multicolumn{4}{c}{\shortstack{Middle tertile\\{\footnotesize($6\%\le\tau/W^\star<20\%$)}}} & \multicolumn{4}{c}{\shortstack{Upper tertile\\{\footnotesize($\tau/W^\star\ge20\%$)}}} \\
\cmidrule(lr){3-6}\cmidrule(lr){7-10}\cmidrule(lr){11-14}
Method & VLM & Self & Cued & Sudden & Overall & Self & Cued & Sudden & Overall & Self & Cued & Sudden & Overall \\
\midrule
Native agent & GPT-6 Astra / high & 11.1 & 61.1 & 60.3 & 44.2 & 16.7 & 37.0 & 38.8 & 30.8 & 0.0 & 8.3 & 29.6 & 12.6 \\
 & GPT-6.1 Sol / xhigh & 48.6 & 57.4 & 44.8 & 50.3 & 27.8 & 40.7 & 53.8 & \textbf{40.8} & 0.0 & 6.9 & 25.5 & 10.8 \\
\addlinespace[2pt]
CaP-X~\cite{capx} & Gemini 3.1 Pro / high$^\dagger$ & 35.2 & 28.7 & 43.3 & 35.7 & 27.8 & 29.6 & 31.2 & 29.5 & 0.0 & 1.4 & 4.3 & 1.9 \\
 & GPT-6.1 Sol / xhigh & 64.4 & 46.8 & 50.3 & 53.8 & 47.2 & 37.5 & 30.7 & 38.5 & 1.4 & 5.6 & 18.7 & 8.5 \\
\addlinespace[2pt]
PyRUA-Lean~\cite{pyrua} & GPT-6 Astra / high$^\dagger$ & 61.1 & 69.0 & 57.9 & \textbf{62.7} & 44.4 & 34.7 & 39.3 & \textbf{39.5} & 22.2 & 18.1 & 24.1 & \textbf{21.5} \\
 & GPT-6.1 Sol / xhigh & 70.4 & 46.3 & 39.0 & 51.9 & 44.4 & 37.5 & 25.9 & 36.0 & 18.1 & 11.1 & 17.6 & \textbf{15.6} \\
\addlinespace[2pt]
Q0-Preview~\cite{q0preview} & Claude Opus 5.5 / xhigh$^\dagger$ & 47.2 & 48.1 & 46.6 & 47.3 & 30.6 & 31.0 & 23.7 & 28.4 & 1.4 & 5.6	 & 14.6 & 7.2 \\
 & GPT-6.1 Sol / xhigh & 63.0 & 63.9 & 56.0 & \textbf{61.0} & 43.1 & 26.9 & 32.5 & 34.1 & 0.0 & 2.8 & 18.6 & 7.1 \\
\bottomrule
\end{tabular*}

\end{table*}

\paragraph{Environment and inputs.}
The experiments run in Isaac Sim with two AgileX PiPER arms. Camera configurations are fixed across time conditions for each method. See the supplementary material for details on method adaptations and the evaluation setup.

\paragraph{Evaluation protocol.}
\label{sec:exp:clock}
Measured-time evaluation examines how task performance responds to reductions in the computation cost of each method--VLM configuration. Controlled-delay evaluation compares methods' sensitivity to a common per-round decision cost.

The measured-time comparison uses $\rho\in\{1,10,100\}$ under the rules of Sec.~\ref{sec:suite:clock}. Table~\ref{tab:exp:latency} reports mean adjusted time per high-level decision round, pooled across both timing conditions. The controlled-delay grid specifies five $\tau$ values per task based on typical window durations, shared across method--VLM configurations; the complete grid is tabulated in the supplementary material. We normalize the task--delay settings by the calibrated reference-window lengths $W^\star$ and group them into tertiles for Table~\ref{tab:exp:controlled}. Each group covers the same task set.

The experiments use 6 scene instances per task. Each instance's object layout, task parameters, and randomized configuration remain fixed across methods, models, and time conditions. For each method, we evaluate two VLM configurations across 18 tasks with six instances per task, yielding 648 measured-time and 1,080 controlled-delay rollouts over the full grids.

\paragraph{Human reference.}
We report human teleoperation performance as a reference for the benchmark. Operators use the head-camera view and control the simulated arms and grippers with two physical arms of the same model, fitted with teaching handles. Throughout a teleoperation rollout, the simulator advances one second for every $\rho$ seconds of real time. The same task judges score human and automated rollouts.

Before evaluation, five operators reviewed task instructions and videos of successful and failed executions, then practiced teleoperation on a subset of tasks to familiarize themselves with the control interface. Human evaluation used three fixed instances per task drawn from the main automated evaluation set, shared across operators and $\rho$ values. Each operator completed one rollout per instance at each $\rho$.

\paragraph{Scoring.}
We report mean task scores as percentages using the episode scores defined in Sec.~\ref{sec:suite:metrics}. For automated methods, we average valid instance scores equally within each task and time condition. Within each normalized-delay group, we average each task's scores across its $\tau$ values. For the human reference, we average operators' scores within each instance before applying the same instance and task weights. Each window-type score weights the tasks of that type equally, and Overall weights the tasks in the comparison equally. Rollouts with infrastructure errors are excluded.

\subsection{Performance under time constraints}
\label{sec:exp:results}

Tables~\ref{tab:main}--\ref{tab:exp:controlled} relate task performance to computation time and decision cost relative to the action window.

\paragraph{Computation time and task completion.}
Automated methods struggle to complete tasks at $\rho=1$ (Table~\ref{tab:main}). Increasing $\rho$ reduces the time consumed by computation while preserving control-step durations, giving the methods more time to complete the required operations. For Native agent with GPT-6.1 Sol/xhigh, the Overall score rises from 0.0\% at $\rho=1$ to 23.3\% at $\rho=10$ and 46.1\% at $\rho=100$.

Zero scores at $\rho=1$ can conceal different degrees of task progress. In inspected Native agent trajectories, an initial interaction can succeed while the remaining manipulation is left incomplete. Traces from other methods also show decision and execution time consuming the window before the required operation is completed. For single-event tasks, intermediate progress receives no credit unless the full success conditions are met. A zero score therefore does not distinguish unsuccessful manipulation from an otherwise useful action sequence that fails to finish in time. The trajectory analysis below examines task progress alongside the time spent on decisions and movements.

Human operators score 85.4\% Overall at $\rho=1$, including 67.2\% on sudden-onset tasks (Table~\ref{tab:main}). Short windows are concentrated in this group: \emph{Catch Ruler} has a mean window of 0.255\,s. Limited reaction and movement time may help explain the lower human score. The supplement reports $W^\star$ for each task.

\paragraph{Performance across relative-delay levels.}
\label{sec:exp:delay}
\label{sec:exp:calibration}

Overall scores are lower in higher normalized-delay groups for every evaluated method--VLM configuration (Table~\ref{tab:exp:controlled}), ranging from 35.7--62.7\% in the lower tertile to 1.9--21.5\% in the upper tertile. Method rankings also change across groups: with the shared GPT-6.1 Sol configuration, Q0-Preview has the highest Overall score in the lower tertile, whereas PyRUA-Lean leads in the upper tertile.

\subsection{Role of window information}
\label{sec:exp:window-information}

At each requested model observation, +Info supplies opening facts and reference time remaining, keeping task dynamics and assigned decision costs unchanged. Table~\ref{tab:exp:window-information} compares Original and +Info scores for Native agent with GPT-6.1 Sol; the recorded rollouts below illustrate how the information enters decisions.

Window information can identify when to leave preparation and whether further observation is affordable. In one \emph{Stamp Paper} rollout, the agent learned that contact with the ink pad had already opened the window. It abandoned further dipping and proceeded to stamp successfully. In a \emph{Heat Milk} rollout with $\tau=5$\,s, the agent cited 7.1\,s remaining, selected immediate shutoff with faster arm motion, and succeeded before closure. Waiting for another observation and then deciding to shut off would incur two 5\,s decisions, exceeding the budget before a new shutoff command could be issued. These cases connect window information to stage recognition and timely action selection, without establishing a fixed benefit ranking across window types.

\begin{table}[t]

\centering
\caption{\textbf{Window information and decision delay.} We compare Original and +Info scores for Native agent using GPT-6.1 Sol. $\Delta$ denotes the difference in percentage points. Scores and differences are computed before rounding to one decimal place.}
\label{tab:exp:window-information}
\small
\setlength{\tabcolsep}{2pt}
\renewcommand{\arraystretch}{1.05}
\begin{tabular*}{\columnwidth}{@{\extracolsep{\fill}}llrrrr@{}}
\toprule
$\tau/W^\star$ & Input & Self & Cued & Sudden & Overall \\
\midrule
$<6\%$ & Original & 48.6 & 57.4 & 44.8 & 50.3 \\
 & +Info & 54.6 & 66.7 & 53.1 & 58.1 \\
 & $\Delta$ & $+6.0$ & $+9.3$ & $+8.3$ & $+7.9$ \\
\midrule
$[6\%,20\%)$ & Original & 27.8 & 40.7 & 53.8 & 40.8 \\
 & +Info & 33.3 & 48.6 & 58.0 & 46.6 \\
 & $\Delta$ & $+5.6$ & $+7.9$ & $+4.2$ & $+5.9$ \\
\midrule
$\ge20\%$ & Original & 0.0 & 6.9 & 25.5 & 10.8 \\
 & +Info & 1.4 & 9.7 & 25.5 & 12.2 \\
 & $\Delta$ & $+1.4$ & $+2.8$ & $+0.0$ & $+1.4$ \\
\bottomrule
\end{tabular*}

\end{table}

\subsection{Decision and execution within action windows}
\label{sec:exp:behavior}

We use selected GPT-6.1 Sol rollouts to examine how observation, decision, and execution timing constrain intervention. Times below are in simulation seconds; the supplementary material provides additional execution details.

\paragraph{Execution per decision varies across methods.}
Per-round latency must be interpreted together with the execution it enables. In one PyRUA-Lean rollout of \emph{Enter Code} at $\rho=10$, a decision charged 4.1\,s produced a program that issued successive control targets over a 1.4\,s execution segment and completed the task before the next decision. Unlike Native agent's direct control-target submissions, PyRUA-Lean can generate a program that issues successive control targets within one high-level decision round. This distinction matters when comparing methods: Table~\ref{tab:exp:latency} measures the cost of a high-level round, while the common per-round charge in Table~\ref{tab:exp:controlled} still permits different amounts of execution between decisions.

\paragraph{Window duration does not determine execution requirements.}
The motion required during an action window depends on the robot configuration at onset. For matched task instances, differences in preparatory actions can lead to different onset configurations and, consequently, different travel and alignment requirements. Normalizing decision cost by $W^\star$ therefore provides a common temporal reference without equalizing physical execution requirements.

\paragraph{Windows can elapse between observations.}
Access to privileged window information remains constrained by observation timing. In an additional CaP-X +Info diagnostic on \emph{Whack-a-mole}, four of six windows opened and closed between observations, leaving their active-window reports unseen by the model. Each decision cost 0.6\,s for a 2.35\,s window, with execution between decisions further extending the observation interval. Privileged window information therefore does not guarantee an opportunity to act on it before closure.

\begin{table}[htbp]
\centering
\caption{{\textbf{Instance score changes from \(\rho=10\) to \(\rho=100\) with GPT-6.1 Sol.}} Entries report the percentages of equally weighted task–instance pairs with valid results under both conditions whose scores improved, remained unchanged, or declined. Percentages are rounded to one decimal place.}
\label{tab:exp:score_changes}
\small
\setlength{\tabcolsep}{3pt}
\begin{tabular*}{\columnwidth}{@{\extracolsep{\fill}}lrrr@{}}
\toprule
Method & Improved & Unchanged & Declined \\
\midrule
Native agent & 36.1 & 47.2 & 16.7 \\
CaP-X~\cite{capx} & 14.8 & 85.2 & 0.0 \\
PyRUA-Lean~\cite{pyrua} & 33.3 & 51.9 & 14.8 \\
Q0-Preview~\cite{q0preview} & 40.7 & 54.6 & 4.6 \\
\bottomrule
\end{tabular*}

\end{table}

\paragraph{Additional execution can offset computational savings.}
Faster computation need not lead to earlier action submission, as both observations and model responses can vary across rollouts. Three of the four methods have instances with declining scores from $\rho=10$ to $100$ (Table~\ref{tab:exp:score_changes}). CaP-X's instance scores at $\rho=10$ are all zero and therefore cannot decline further. Paired Native agent rollouts on the same \emph{Conveyor Picking} instance illustrate this effect: before the first gripper-closing command, the higher-$\rho$ rollout saved 3.2\,s in computation but spent 4.4\,s longer executing between decisions. Additional observations and arm repositioning consumed more time than faster computation saved, delaying submission of the gripper-closing command.

\section{Conclusion}
\label{sec:conclusion}

AWAIT evaluates manipulation under process-dependent action windows and groups tasks by how information about window opening becomes available.
Visual cues and scoring share the same process state, while decision time and physical execution are accounted for on a common simulation clock throughout each task episode.

In the measured-time comparison, increasing $\rho$ from 10 to 100 raises the Overall score for every evaluated method--VLM configuration on fixed task instances.
For Native agent with GPT-6.1 Sol, explicit window-opening and remaining-time information improves Overall scores across the evaluated relative-delay groups, with inspected traces showing how the reports guide stage recognition and action selection.
Paired execution traces also show that time saved in computation can be offset by additional observations and movements, motivating evaluation of the complete intervention sequence alongside per-round decision cost.
AWAIT thus supports comparisons of systems' sensitivity to decision costs.

\clearpage

{
    \small
    \bibliographystyle{ieeenat_fullname}
    \bibliography{main}

\begin{thebibliography}{32}
\providecommand{\natexlab}[1]{#1}
\providecommand{\url}[1]{\texttt{#1}}
\expandafter\ifx\csname urlstyle\endcsname\relax
  \providecommand{\doi}[1]{doi: #1}\else
  \providecommand{\doi}{doi: \begingroup \urlstyle{rm}\Url}\fi

\bibitem[Cao et~al.(2025)Cao, Hong, Chen, Pan, and Liu]{physxanything}
Ziang Cao, Fangzhou Hong, Zhaoxi Chen, Liang Pan, and Ziwei Liu.
\newblock {PhysX-Anything}: Simulation-ready physical {3D} assets from single
  image, 2025.

\bibitem[Chen et~al.(2026{\natexlab{a}})Chen, Chen, Li, Tang, Su, Lu,
  et~al.]{robodojo}
Tianxing Chen, Yue Chen, Zixuan Li, Junyuan Tang, Kailun Su, Haoran Lu, et~al.
\newblock {RoboDojo}: A unified sim-and-real benchmark for comprehensive
  evaluation of generalist robot manipulation policies, 2026{\natexlab{a}}.

\bibitem[Chen et~al.(2026{\natexlab{b}})Chen, Zhang, and Li]{reflexbench}
Yuxuan Chen, Wanruo Zhang, and Xiao Li.
\newblock {Reflex}: Enabling fast and predictive vision-language-action models
  for reaction-critical manipulation, 2026{\natexlab{b}}.
\newblock Introduces the ReflexBench benchmark for reaction-critical
  manipulation.

\bibitem[Fang et~al.(2026)Fang, Li, Wang, Xi, Liang, and Bai]{domino}
Heng Fang, Shangru Li, Shuhan Wang, Xuanyang Xi, Dingkang Liang, and Xiang Bai.
\newblock Towards generalizable robotic manipulation in dynamic environments.
\newblock In \emph{European Conference on Computer Vision (ECCV)}, 2026.

\bibitem[Fu et~al.(2026)Fu, Yu, El-Refai, Kou, Xue, et~al.]{capx}
Letian Fu, Justin Yu, Karim El-Refai, Ethan Kou, Haoru Xue, et~al.
\newblock {CaP-X}: A framework for benchmarking and improving coding agents for
  robot manipulation, 2026.

\bibitem[Gong et~al.(2023)Gong, Huang, Zhao, Geng, Gao, Wu, Ai, Zhou,
  Terzopoulos, Zhu, Jia, and Huang]{arnold}
Ran Gong, Jiangyong Huang, Yizhou Zhao, Haoran Geng, Xiaofeng Gao, Qingyang Wu,
  Wensi Ai, Ziheng Zhou, Demetri Terzopoulos, Song-Chun Zhu, Baoxiong Jia, and
  Siyuan Huang.
\newblock {ARNOLD}: A benchmark for language-grounded task learning with
  continuous states in realistic {3D} scenes.
\newblock In \emph{Proceedings of the IEEE/CVF International Conference on
  Computer Vision (ICCV)}, pages 20483--20495, 2023.

\bibitem[Gonzalez-Pumariega et~al.(2025)Gonzalez-Pumariega, Yean, Sunkara, and
  Choudhury]{robotouille}
Gonzalo Gonzalez-Pumariega, Leong~Su Yean, Neha Sunkara, and Sanjiban
  Choudhury.
\newblock {Robotouille}: An asynchronous planning benchmark for {LLM} agents,
  2025.

\bibitem[Huang et~al.(2023)Huang, Wang, Zhang, Li, Wu, and Fei-Fei]{voxposer}
Wenlong Huang, Chen Wang, Ruohan Zhang, Yunzhu Li, Jiajun Wu, and Li Fei-Fei.
\newblock {VoxPoser}: Composable {3D} value maps for robotic manipulation with
  language models.
\newblock In \emph{Proceedings of The 7th Conference on Robot Learning}, pages
  540--562. PMLR, 2023.

\bibitem[Ichter et~al.(2023)Ichter, Brohan, Chebotar, Finn, Hausman, Herzog,
  et~al.]{saycan}
Brian Ichter, Anthony Brohan, Yevgen Chebotar, Chelsea Finn, Karol Hausman,
  Alexander Herzog, et~al.
\newblock Do as i can, not as i say: Grounding language in robotic affordances.
\newblock In \emph{Proceedings of The 6th Conference on Robot Learning}, pages
  287--318. PMLR, 2023.

\bibitem[Jia et~al.(2025)Jia, Yan, Hao, Li, Zhang, He, Li, Wang, Chen, Jiang,
  Yin, Quan, Chen, and Yuan]{ultrashape}
Tanghui Jia, Dongyu Yan, Dehao Hao, Yang Li, Kaiyi Zhang, Xianyi He, Lanjiong
  Li, Yuhan Wang, Jinnan Chen, Lutao Jiang, Qishen Yin, Long Quan, Ying-Cong
  Chen, and Li Yuan.
\newblock {UltraShape 1.0}: High-fidelity {3D} shape generation via scalable
  geometric refinement, 2025.

\bibitem[Joshi et~al.(2025)Joshi, Han, Nugent, Saez-Diez, Zuo,
  et~al.]{infinigenarticulated}
Abhishek Joshi, Beining Han, Jack Nugent, Max~Gonzalez Saez-Diez, Yiming Zuo,
  et~al.
\newblock Procedural generation of articulated simulation-ready assets, 2025.

\bibitem[Lee et~al.(2026)Lee, Baek, Kim, Jeon, Kim, Kim, and Kwak]{robotuse}
Junhoo Lee, Injun Baek, Seungyeon Kim, Suhyun Jeon, Minkyu Kim, Baekseung Kim,
  and Nojun Kwak.
\newblock {RobotUse}: Allocating computation, context, and decisions, 2026.

\bibitem[Li et~al.(2024{\natexlab{a}})Li, Zhang, Wong, Gokmen, Srivastava,
  et~al.]{behavior1k}
Chengshu Li, Ruohan Zhang, Josiah Wong, Cem Gokmen, Sanjana Srivastava, et~al.
\newblock {BEHAVIOR-1K}: A human-centered, embodied {AI} benchmark with 1,000
  everyday activities and realistic simulation, 2024{\natexlab{a}}.

\bibitem[Li et~al.(2026{\natexlab{a}})Li, Ju, Qin, Fang, and Li]{homeworld}
Wenbo Li, Xiaoliang Ju, Zipeng Qin, Rongyao Fang, and Hongsheng Li.
\newblock {HomeWorld}: A unified floorplan-to-furnished framework for
  generating controllable, densely interactive whole-home scenes,
  2026{\natexlab{a}}.

\bibitem[Li et~al.(2024{\natexlab{b}})Li, Li, Zhang, Zhang, Jia,
  et~al.]{robogsim}
Xinhai Li, Jialin Li, Ziheng Zhang, Rui Zhang, Fan Jia, et~al.
\newblock {RoboGSim}: A {Real2Sim2Real} robotic gaussian splatting simulator,
  2024{\natexlab{b}}.

\bibitem[Li et~al.(2026{\natexlab{b}})Li, Yang, Zhao, and Rasouli]{robodyna}
Zhiyuan Li, Ruiheng Yang, Xuan Zhao, and Amir Rasouli.
\newblock Benchmarking robotic manipulation under physically realistic
  dynamics.
\newblock In \emph{NeurIPS 2026 Workshop on Robot Learning with World Models},
  2026{\natexlab{b}}.

\bibitem[Liang et~al.(2022)Liang, Huang, Xia, Xu, Hausman, Ichter, Florence,
  and Zeng]{cap}
Jacky Liang, Wenlong Huang, Fei Xia, Peng Xu, Karol Hausman, Brian Ichter, Pete
  Florence, and Andy Zeng.
\newblock Code as policies: Language model programs for embodied control, 2022.

\bibitem[Lin et~al.(2021)Lin, Wang, Olkin, and Held]{softgym}
Xingyu Lin, Yufei Wang, Jake Olkin, and David Held.
\newblock {SoftGym}: Benchmarking deep reinforcement learning for deformable
  object manipulation.
\newblock In \emph{Proceedings of the 2020 Conference on Robot Learning}, pages
  432--448. PMLR, 2021.

\bibitem[Liu et~al.(2026)Liu, Mei, Xian, Han, Sun, Chen, et~al.]{motionforge}
Mohan Liu, Dengchen Mei, Haotian Xian, Ruyang Han, Jiayi Sun, Xuanyu Chen,
  et~al.
\newblock {MotionForge}: A data generation pipeline and large-scale benchmark
  for long-horizon manipulation of dynamic objects with domain shifts, 2026.

\bibitem[{MatWeb, LLC}()]{matweb}
{MatWeb, LLC}.
\newblock {MatWeb}: Online materials information resource.
\newblock \url{https://www.matweb.com/}.
\newblock Online material property database.

\bibitem[{NVIDIA et al.}(2026)]{gr00t2}
{NVIDIA et al.}
\newblock {GR00T 2}: Scaling robot foundation models for generalization,
  dexterity, and adaptation.
\newblock Whitepaper, NVIDIA, 2026.

\bibitem[Pednekar et~al.(2026)Pednekar, Garrido-Perez, Khaluf, and
  Simoens]{chronocooked}
Amrapali Pednekar, Alvaro Garrido-Perez, Yara Khaluf, and Pieter Simoens.
\newblock {Chronocooked}: A benchmark for interval timing in reinforcement
  learning agents, 2026.

\bibitem[Qiu et~al.(2026)Qiu, Lei, Huang, and Ying]{dysta}
Weikang Qiu, Huashuo Lei, Tinglin Huang, and Rex Ying.
\newblock Efficient long-horizon vision-language-action models via
  static-dynamic disentanglement, 2026.

\bibitem[{Qoo Tech}(2026)]{q0preview}
{Qoo Tech}.
\newblock {Q0-Preview}.
\newblock Software implementation in XPolicyLab, 2026.
\newblock Accessed November 10, 2026.

\bibitem[Qureshi et~al.(2024)Qureshi, Garg, Yandun, Held, Kantor, and
  Silwal]{splatsim}
Mohammad~Nomaan Qureshi, Sparsh Garg, Francisco Yandun, David Held, George
  Kantor, and Abhisesh Silwal.
\newblock {SplatSim}: Zero-shot {Sim2Real} transfer of {RGB} manipulation
  policies using gaussian splatting, 2024.

\bibitem[Ren et~al.(2026)Ren, Sun, Liu, Chen, Jin, Zou, and
  Chen]{liberodynamic}
Shaojie Ren, Jingwen Sun, Zezhi Liu, Hongjin Chen, Xin Jin, Wei Zou, and Zhibo
  Chen.
\newblock {LIBERO-Dynamic}: Benchmarking robotic manipulation in dynamic
  environments, 2026.

\bibitem[Si et~al.(2026)Si, Bi, Yang, Ni, Huang, Wang, Jiang, Zhang, Ruan,
  Zhang, Wang, Li, Feng, Dong, and Zhou]{pyrua}
Ruiyang Si, Jianxin Bi, Shunyu Yang, Rui Ni, Wenbo Huang, Qiang Wang, Shulong
  Jiang, Yintianrun Zhang, Feixiang Ruan, Kewei Zhang, Duomin Wang, Xiuyu Li,
  Haiwen Feng, Zhen Dong, and Daquan Zhou.
\newblock Fewer tokens, better action: {GPT-6 Astra} robot agents with 14\%
  higher success rate but 65\% fewer tokens, 2026.

\bibitem[Sung et~al.(2026)Sung, Leijiahao, Wang, Tan, Fu, Tan, Li, Chen, Wan,
  Huang, Liang, and Hua]{liberoindustry}
Ming Shin~Cindy Sung, Duan Leijiahao, Haoxiang Wang, Boning Tan, Liangna Fu,
  Chao Tan, Yafu Li, Yulong Chen, Chaoqun Wan, Zhen Huang, Shuxian Liang, and
  Xian-Sheng Hua.
\newblock {LIBERO-Industry}: Benchmarking robotic manipulation under industrial
  process constraints, 2026.

\bibitem[Xian et~al.(2023)Xian, Zhu, Xu, Tung, Torralba, Fragkiadaki, and
  Gan]{fluidlab}
Zhou Xian, Bo Zhu, Zhenjia Xu, Hsiao-Yu Tung, Antonio Torralba, Katerina
  Fragkiadaki, and Chuang Gan.
\newblock {FluidLab}: A differentiable environment for benchmarking complex
  fluid manipulation.
\newblock In \emph{International Conference on Learning Representations
  (ICLR)}, 2023.

\bibitem[Xiang et~al.(2025)Xiang, Chen, Xu, Wang, Lv, Deng, Zhu, Dong, Zhao,
  Yuan, and Yang]{trellis2}
Jianfeng Xiang, Xiaoxue Chen, Sicheng Xu, Ruicheng Wang, Zelong Lv, Yu Deng,
  Hongyuan Zhu, Yue Dong, Hao Zhao, Nicholas~Jing Yuan, and Jiaolong Yang.
\newblock Native and compact structured latents for {3D} generation, 2025.

\bibitem[Xie et~al.(2026)Xie, Wen, Zheng, Chen, Hong, Diao, and
  Liu]{dynamicvla}
Haozhe Xie, Beichen Wen, Jiarui Zheng, Zhaoxi Chen, Fangzhou Hong, Haiwen Diao,
  and Ziwei Liu.
\newblock {DynamicVLA}: A vision-language-action model for dynamic object
  manipulation, 2026.
\newblock NeurIPS 2026.

\bibitem[Zook et~al.(2025)Zook, Sun, Spjut, Blukis, Birchfield, and
  Tremblay]{grs}
Alex Zook, Fan-Yun Sun, Josef Spjut, Valts Blukis, Stan Birchfield, and
  Jonathan Tremblay.
\newblock {GRS}: Generating robotic simulation tasks from real-world images,
  2025.

\end{thebibliography}
}

\twocolumn[{\centering\Large\bfseries Appendix\par\vspace{1em}}]
\appendix
\EnableAppendixBottomFloats
\section{Additional Results}
\label{sec:supp:additional-results}

\subsection{Evaluation across scene domains}
\label{sec:supp:scene-domains}

\begin{table*}[!b]
\begingroup
\centering
\caption{\textbf{Controlled delay across scene domains.} Mean task scores (\%) across factory, caf\'e, and kindergarten scenes for GPT-6.1 Sol / xhigh. Within each tertile, we average scores over scenes and delays for each task, then average tasks with equal weights. The best two Overall scores per tertile are in bold.}
\label{tab:supp:scene-domains}
\small
\renewcommand{\arraystretch}{1.10}
\setlength{\tabcolsep}{1.8pt}
\begin{tabular*}{\textwidth}{@{\extracolsep{\fill}}l*{12}{r}@{}}
\toprule
& \multicolumn{4}{c}{\shortstack{Lower tertile\\{\footnotesize($\tau/W^\star<6\%$)}}} & \multicolumn{4}{c}{\shortstack{Middle tertile\\{\footnotesize($6\%\le\tau/W^\star<20\%$)}}} & \multicolumn{4}{c}{\shortstack{Upper tertile\\{\footnotesize($\tau/W^\star\ge20\%$)}}} \\
\cmidrule(lr){2-5}\cmidrule(lr){6-9}\cmidrule(lr){10-13}
Method & Self & Cued & Sudden & Overall & Self & Cued & Sudden & Overall & Self & Cued & Sudden & Overall \\
\midrule
Native agent & 48.1 & 58.3 & 45.6 & 50.7 & 27.8 & 41.7 & 52.7 & \textbf{40.7} & 0.0 & 5.6 & 25.9 & \textbf{10.5} \\
CaP-X & 64.8 & 47.2 & 51.4 & \textbf{54.5} & 47.2 & 37.0 & 30.1 & \textbf{38.1} & 2.8 & 5.6 & 18.9 & 9.1 \\
PyRUA-Lean & 67.6 & 46.3 & 39.8 & 51.2 & 44.4 & 38.0 & 26.3 & 36.2 & 16.7 & 11.1 & 17.4 & \textbf{15.0} \\
Q0-Preview & 64.8 & 60.2 & 54.9 & \textbf{60.0} & 41.7 & 26.9 & 32.6 & 33.7 & 0.0 & 2.8 & 18.1 & 6.9 \\
\bottomrule
\end{tabular*}
\par
\endgroup

\end{table*}

To isolate the effects of decision timing and computation cost from visual clutter, the main comparisons in the paper hold the surrounding environment fixed to a uniform plain background. In practice, however, embodied manipulation often takes place in visually complex settings where clutter, lighting, and surrounding geometry can make task cues harder to identify, and extra observations or corrective motions leave less time for intervention. Here, we evaluate the methods' robustness to changes in the visual background and examine whether their scores still decline as normalized delay increases.

For each of the 18 tasks, we place the robot and task objects in three photorealistic 3D environments generated by our pipeline: a factory, a caf\'e, and a kindergarten. We adapt these placements to each scene while holding task dynamics and the parameters governing action-window duration fixed. We check visibility and reachability and verify that the task can be completed in each instance. All four methods use GPT-6.1 Sol / xhigh.

\paragraph{Aggregate performance across scenes.}
After pooling the three scenes, each method's Overall score remains within about one percentage point of its corresponding uniform-background score in every normalized-delay tertile, and all four methods continue to score progressively lower from the lowest to the highest delay tertile (Table~\ref{tab:supp:scene-domains}). These results indicate that varying the surrounding scene domain has minimal impact on aggregate task performance and preserves the delay-sensitivity trends observed in the main evaluation.

\paragraph{Agents can distinguish task targets from background distractors.}
A CaP-X rollout of \emph{Pour Liquid} in the factory ($\tau=0.2$\,s) illustrates a pattern we observed across methods. Rows of bottles on the rear conveyors resembled the foreground task bottle. SAM3 returned masks for the task bottle and for objects on the conveyors. The agent excluded the conveyor bottles in its point queries and deferred grasping. It then selected the highest-scoring mask, which contained the query point on the task bottle's red label, and issued a grasp command for that bottle.

\subsection{Per-task controlled-delay scores}
\label{sec:supp:per-task-results}

Table~\ref{tab:supp:per-task-tau} reports scores for all 18 tasks under controlled decision delays, marginalized over all five delays for each method and over all four methods for each normalized-delay tertile. Aggregating across methods confirms that every task exhibits a monotonic score decline from the lower to the upper delay tertile, rather than the aggregate trend being driven by a subset of tasks. Across methods, per-task scores remain well below saturation on most tasks, leaving substantial headroom across all three window types.

\begin{table*}[t]
\centering
\small
\setlength{\tabcolsep}{5pt}
\caption{\textbf{Per-task scores under controlled decision delays.} Mean task scores (\%) for GPT-6.1 Sol / xhigh. Left: scores for each method averaged over all five controlled delays. Right: scores for each normalized-delay tertile (Lower: $\tau/W^\star < 6\%$; Middle: $6\% \le \tau/W^\star < 20\%$; Upper: $\tau/W^\star \ge 20\%$) averaged over all four methods.}
\label{tab:supp:per-task-tau}
\begin{tabular*}{\textwidth}{@{\extracolsep{\fill}}l*{4}{r}*{3}{r}@{}}
\toprule
 & \multicolumn{4}{c}{By method (all delays)} & \multicolumn{3}{c}{By delay tertile (all methods)} \\
\cmidrule(lr){2-5}\cmidrule(lr){6-8}
Task & Native agent & CaP-X & PyRUA-Lean & Q0-Preview & Lower & Middle & Upper \\
\midrule
\multicolumn{8}{l}{\emph{Self-triggered windows}} \\
Withdraw Cash & 20.0 & 33.3 & 36.7 & 26.7 & 62.5 & 35.4 & 6.2 \\
Glue Handle & 36.7 & 46.7 & 53.3 & 43.3 & 59.7 & 41.7 & 4.2 \\
Enter Code & 36.7 & 50.0 & 56.7 & 50.0 & 63.9 & 45.8 & 4.2 \\
Clean Mirror & 20.0 & 33.3 & 46.7 & 33.3 & 58.3 & 41.7 & 4.2 \\
Stamp Paper & 20.0 & 30.0 & 36.7 & 30.0 & 62.5 & 37.5 & 4.2 \\
Blow Bubbles & 26.7 & 36.7 & 46.7 & 33.3 & 62.5 & 41.7 & 6.2 \\
\midrule
\multicolumn{8}{l}{\emph{Cued windows}} \\
Inflate Balloon & 30.0 & 26.7 & 30.0 & 23.3 & 54.2 & 37.5 & 4.2 \\
Dispense Water & 33.3 & 30.0 & 26.7 & 33.3 & 52.1 & 33.3 & 8.3 \\
Flip Hourglass & 30.0 & 30.0 & 26.7 & 26.7 & 54.2 & 37.5 & 6.2 \\
Conveyor Picking & 43.3 & 30.0 & 36.7 & 43.3 & 52.8 & 29.2 & 4.2 \\
Heat Milk & 40.0 & 30.0 & 33.3 & 30.0 & 58.3 & 34.7 & 4.2 \\
Pour Liquid & 46.7 & 36.7 & 36.7 & 36.7 & 50.0 & 41.7 & 12.5 \\
\midrule
\multicolumn{8}{l}{\emph{Sudden-onset windows}} \\
Stop Dominoes & 43.3 & 26.7 & 23.3 & 30.0 & 45.8 & 35.4 & 18.8 \\
Whack-a-mole & 38.3 & 30.6 & 25.0 & 32.8 & 49.3 & 33.7 & 20.8 \\
Catch Ruler & 43.3 & 30.0 & 23.3 & 30.0 & 45.8 & 37.5 & 18.8 \\
Piano Tiles & 39.2 & 31.7 & 26.2 & 33.8 & 48.4 & 34.9 & 22.7 \\
Catch Cup & 36.7 & 33.3 & 30.0 & 36.7 & 47.9 & 37.5 & 18.8 \\
Goalkeeping & 43.3 & 36.7 & 30.0 & 40.0 & 47.9 & 35.4 & 20.8 \\
\bottomrule
\end{tabular*}
\end{table*}
\section{Evaluation Protocol and Method Integration}
\label{sec:exp:integration}

\subsection{Evaluation platform, black-box interface, and isolation}
\label{sec:supp:platform}

\paragraph{Simulation and initialization.}
The host runs automated rollouts in Isaac Sim~6.0.1 with PhysX~5 (extension version \texttt{110.1.13}) at a standard physics rate of 240\,Hz. Local perception and planning tools run on a workstation with an Intel Xeon Silver~4310 CPU and two NVIDIA GeForce RTX~4090~D GPUs (24GB each). The default scene solver is temporal Gauss--Seidel (TGS), with object-specific solver settings stored in the USD assets. The four tasks with particle liquids (\emph{Pour Liquid}, \emph{Clean Mirror}, \emph{Heat Milk}, and \emph{Dispense Water}) use GPU dynamics and a GPU broad phase, with 12 particle position iterations per step and at most 96 neighbors per particle. Asset loading, physics-handle creation, and connecting the robot controller, task mechanisms, and judge all complete before the first physics step. The initial observation is rendered at simulation time zero without advancing physics, and time accounting, execution, and scoring start from that state.

\paragraph{Physics-step execution.}
On every physics step, the host applies the active drive targets, advances PhysX, updates the task mechanisms, and queries the judge. Environment dynamics and task evaluation therefore operate at the simulation step rate rather than at the agent's observation or decision frequency, and continue without interruption between observations and throughout intervals charged to computation.

\paragraph{Robot models and control.}
The host uses a shared articulation controller with model-specific USD assets, joint mappings, drive parameters, and gripper calibration. The controller checks command limits and converts accepted targets into position drive references at each physics step. In the reported experiments, all methods use the same two PiPER arms. Each arm caps the speed of its first five joints at 5\,rad/s and the sixth at 3\,rad/s. The arm joint position drives use a stiffness of 100\,N\,m/rad, a damping coefficient of 10\,N\,m\,s/rad, and a torque limit of 100\,N\,m. The gripper opening ranges from 0 to 70\,mm, with a force limit of 10\,N per finger. The PiPER controller accepts joint position targets, calibrated gripper opening and closing preload, and straight-line tool motion using the PiPER kinematic model.

\paragraph{Public observations and feedback.}
The rig contains one head camera and one camera attached to each wrist. Each observation pairs the camera images and their current transforms with measured robot state at the same simulation time, without advancing physics. Public camera images have a resolution of $640\times480$ pixels. Each view includes camera intrinsics and a camera-to-world transform in the OpenCV optical convention. Where provided, depth is measured along the optical axis in meters. Camera mounting and intrinsics are fixed for each instance across methods, models, and timing conditions; the wrist camera transforms follow the arms. Robot proprioception comprises measured joint positions and velocities, TCP poses, and gripper opening, together with joint and actuator limits. Active command targets and gripper preload values are also reported as control settings, without contact-force or grasp-success measurements.

\paragraph{Action feedback and replay.}
The host reports which commands are pending, active, replaced, or rejected. It also reports tracking errors relative to the active joint, tool-pose, and gripper targets. For each accepted command, the action trace records when it was submitted and when it took effect, if it started. To replay a rollout, we load the same initial instance and apply the commands at their recorded activation times without rerunning the policy.

\paragraph{Black-box interface and process isolation.}
The host evaluates each method as a black box through a trusted adapter and manages time accounting, simulation, actuator limits, and task judging. Each method runs in a separate process from the Isaac Sim host and interacts with the environment only through this public interface, isolating it from host task code, privileged USD object poses, and scoring state. The adapter maps the method's environment interactions to four core primitives: \texttt{observe} reads public observations and robot feedback, \texttt{advance} submits control targets, \texttt{decision} marks high-level decision round boundaries, and \texttt{elapsed} accounts for time and advances simulation. Of these primitives, only \texttt{elapsed} advances physics; \texttt{advance} queues a target for a subsequent physics step. A control step submits a target when needed, advances simulation through \texttt{elapsed} for the requested duration, and returns robot feedback. The host also uses \texttt{elapsed} to advance simulation for accounted computation or a controlled-delay charge, with the robot tracking the previously active targets.

\subsection{Unified clock and computation-time accounting}
\label{sec:supp:time-accounting}
\label{sec:supp:decision-granularity}

\paragraph{High-level decision rounds.}
A high-level decision round groups the computations used to decide what the robot should do next and can include several model requests and tool calls. A running program can read local feedback and submit further control targets without opening a new round.

\paragraph{Execution and timing.}
The host settles each round's charge before accepting its first new target. Each method retains the control intervals of its execution loop, and each interval can span multiple physics steps. In measured time, tool computation between steps also advances simulation under the active targets. The computation meter excludes simulator service time, including physics advancement and observation rendering; those operations are not charged again as policy computation.

The main VLMs are served remotely. Model initialization and reusable dependency loading take place before task input is delivered. Measured computation includes input-dependent perception, code execution, and motion planning, as well as time in local service and model-request admission queues.

Under controlled delay, the host advances simulation by $\tau$ once per high-level round, at its close, while the robot tracks its previous targets. Computation inside the resulting execution routine has no additional simulation-time charge, but every requested physical control interval still advances time. In the measured-time condition, both high-level computation and computation within execution routines advance simulation by their adjusted real duration divided by $\rho$. This scaling leaves joint and gripper speed limits unchanged. Closing the round does not charge its computation a second time. High-level rounds, model requests, submitted targets, and physics steps are therefore distinct counts.

Targets take effect on a subsequent physics step. A new command replaces the corresponding active targets; omitted arm fields retain their previous values. A wait does not cancel motion or reset a task process. If a method finishes early, the host keeps advancing simulation and evaluating the task under the last targets until the episode horizon. The task judge determines success, process failure, and time limits independently of the method's completion declaration.

\paragraph{First-event wait adjustment.}
We identify deductible response waits for both model access routes:
\begin{itemize}
  \item \textbf{Codex subscription.} For GPT requests, the wait begins immediately before the HTTP send and ends when the Codex transport parses the first nonempty JSON event in the response stream.
  \item \textbf{Streaming API access.} For Claude and Gemini requests, streaming API calls use the same send-to-first-event boundary on the provider's response stream.
\end{itemize}
Queueing before the send and computation after the first event remain chargeable. Requests without a recorded first event are timed in full.

For concurrent workers, let $\mathcal{A}(u)$ be the active computation workers at real time $u$ and $\mathcal{Q}(u)$ those in confirmed deductible response waits. The adjusted duration over a measured interval $[a,b]$ is
\begin{equation}
\Delta t^{\mathrm{real}}=\int_a^b
\mathbf{1}\!\left[\mathcal{A}(u)\setminus\mathcal{Q}(u)\ne\varnothing\right]du.
\label{eq:supp:time-union}
\end{equation}
This union accounting retains computation that overlaps another worker's wait and counts concurrent computation once. It applies to CaP-X's parallel candidates and Q0-Preview's planning and execution workers. It is elapsed system time after the stated adjustment, not a sum of worker CPU times or a hardware-independent measure of model computation.

\paragraph{Example: \emph{Enter Code}.}
Table~\ref{tab:supp:round-trace} traces three high-level rounds in a PyRUA-Lean rollout. The window opened at 2.69\,s while the program from round~1 was running. The agent captured its first high-level observation after the window opened at the start of round~2. The program from that round submitted 33 control targets. The task succeeded during this sequence at 8.70\,s, before the agent captured the next observation. In PyRUA-Lean, the program reads robot feedback directly during execution, while tool results and any returned images enter the conversation history for the next model request.

\begin{table}[htbp]
\centering
\footnotesize
\caption{\textbf{High-level rounds in \emph{Enter Code}.} This is the rollout discussed in Sec.~4.4 of the main paper (PyRUA-Lean with GPT-6.1 Sol / xhigh, $\rho=10$). Capture is the observation time; Send and Return mark the first model request and the last response in each round. Times are in simulation seconds. The Targets column counts control target submissions, including interpolated targets.}
\label{tab:supp:round-trace}
\begin{tabular*}{\columnwidth}{@{\extracolsep{\fill}}rrrrr@{}}
\toprule
Round & Capture & Send & Return & Targets \\
\midrule
1 & 0.000 & 0.083 & 1.675  & 32 \\
2 & 3.754 & 3.792 & 7.904  & 33 \\
3 & 9.317 & 9.354 & 10.408 & 0 \\
\bottomrule
\end{tabular*}
\end{table}

\FloatBarrier
\subsection{Adapter implementations for the four methods}
\label{sec:supp:adapters}

\paragraph{Native agent.}
Native agent receives only the head view and measured robot state. We restrict it to a single camera to keep it a minimal, code-free VLM baseline aligned with the head-only human teleoperation setup, and to avoid inflating per-turn visual context and inference latency during direct multi-turn control. The adapter encodes depth as a grayscale image aligned with the RGB pixels on a fixed 0--5\,m scale (0 for invalid depth and 1--255 for valid depth). The method uses public robot kinematics to specify arm and gripper targets. Each observation starts a high-level decision round, which closes before a valid target submission, on an explicit wait, or at rollout termination. Target validation, inverse kinematics, and corrections before an accepted command remain within the same round. The adapter executes the targets through the public robot controller, without a code-execution environment or a task-specific planner. An explicit wait advances the control interval under the existing targets before the next observation.

\paragraph{CaP-X.}
We integrate CaP-X at revision \texttt{53e9966}, retaining its nine parallel program candidates, subsequent synthesis, multi-turn visual differencing, and perception and control skill library. The method receives RGBD observations from the head camera and both wrist cameras, with numerical depth for its perception tools. It retains SAM3 segmentation at revision \texttt{6fe87d6}, Contact-GraspNet proposals, PyRoKi inverse kinematics, and the Molmo2-8B image-pointing tool. Initial visual feedback, code generation, or multi-turn replanning opens a decision round when none is active. Candidate generation and synthesis belong to the same round and incur a single controlled-delay charge. The first control step, successful program return, or rollout termination closes the round. A failed program that has not executed a physical step can be corrected within the open round; a failure after execution does not undo an already charged round. Feedback reads during program execution can guide subsequent control steps without an additional high-level decision charge.

\paragraph{PyRUA-Lean.}
We integrate revision \texttt{e754cf4} with RGBD observations from the head camera and both wrist cameras, retaining numerical depth for perception. The adapter back-projects the initial head-camera depth to estimate the dominant horizontal support surface near the robot bases, then uses its height to align camera and robot poses to the native planning frame. The adapter retains the Python runtime, SAM3 segmentation at revision \texttt{c1f6b1e}, native inverse kinematics, and closed-loop motion and gripper routines, without a vision--language--action policy. The initial observation begins the first decision round; a subsequent model request opens another round when none is active. The first control step or rollout termination closes the round. Local feedback within a motion routine remains part of the same intervention.

\paragraph{Q0-Preview.}
We integrate revision \texttt{6baccd5}, preserving concurrent planning and execution, code tools, and CuRobo at revision \texttt{895c651} for motion planning and execution checks. Consistent with Q0-Preview's upstream design, its generated Python code executes in an isolated sandboxed runtime (implemented with an unprivileged \texttt{chroot} backend) that exposes only public dependencies, robot resources, and the episode's public observations while preserving its upstream planning, code-tool, and execution-check structure. Q0-Preview's model receives RGB images from the head camera and both wrist cameras, together with camera calibration. Its adapter uses the same depth-based support plane estimation as PyRUA-Lean to align the planning frame. The initial observation starts the first decision round, and a planning observation opens a later round when none is active. The next control step or rollout termination closes the round; execution checks can proceed within the same intervention.

\subsection{Aggregation and diagnostic protocols}
\label{sec:supp:evaluation-coverage}

\paragraph{Score weights and exclusions.}
For each condition, we average valid automated instance scores within each task, giving equal weight to all instances. Task-type scores and the Overall score give equal weight to the corresponding tasks. For teleoperation, we first average scores across operators for each instance. Normal zero scores and failures before window opening remain in the score denominator; only infrastructure failures are excluded. Across all our experiments, infrastructure exclusions arose exclusively from transient remote-API timeouts during periods of upstream service congestion rather than simulator crashes or method-specific faults. To handle intermittent network and backend instability without corrupting episode timing or scores, the evaluation harness automatically retries timed-out model requests up to five times; if a request still fails after five consecutive retries, the harness aborts the episode and discards the partial rollout. Such aborts occurred in at most 3\% of rollouts and reflect external API availability rather than task difficulty or method behavior.

\paragraph{Latency weights.}
Decision-latency statistics pool the adjusted elapsed real times of high-level decision rounds from both timing conditions under GPT-6.1 Sol/xhigh. The mean is the sum of those durations divided by the number of rounds, rather than an equal-weight average of task or episode means. A round can contain several model calls and tool computations. Computation performed only within an executing program contributes to measured-time advancement but is not a separate high-level round. Under controlled delay, the adjusted elapsed real time remains a latency measurement, while the assigned simulation-time cost is $\tau$.

\paragraph{Window information (+Info).}
In +Info, the host attaches a window snapshot to each requested observation, and the adapter includes it in the model input. The snapshot records the simulation time of the observation and lists the windows that have already opened, in opening order. For each opened window, it reports whether the reference deadline $t_{\mathrm{close}}$, defined by the task's prescribed continuation without further intervention, is still ahead or has elapsed, along with the remaining simulation time (zero at or after the deadline). The snapshot updates only at observation boundaries, so opening a window does not trigger a new model observation. Original uses the same task dynamics, judge, and timing rules without the window snapshot.

\subsection{Native agent prompts and +Info input}
\label{sec:supp:prompt-examples}

\paragraph{Prompt template.}
The Native agent receives a system message with control instructions, a fixed user message with the task prompt from the task catalog and the robot and camera specifications, and a user message for each observation. The latter starts with \texttt{Observation (JSON):} and contains the measured state and action feedback, followed by the head-camera RGB and depth images. Figure~\ref{fig:supp:native-prompt} summarizes the system prompt's main control instructions and shows an example \texttt{+Info} observation snippet.

\paragraph{Tools.}
\texttt{submit} takes a required JSON object \texttt{command}, keyed by \texttt{left} and/or \texttt{right}. Each arm entry can specify a motion target, speed limits, gripper opening, and closing preload. \texttt{wait} takes no command and advances to the next observation under the active targets. Both tools accept an optional \texttt{memory} string of up to 2,400 characters; neither accepts other arguments.

\paragraph{+Info observation example.}
In +Info, the fixed message defines the window fields, and the observation JSON includes a \texttt{window\_information} object. The excerpt in Figure~\ref{fig:supp:native-prompt} uses the values at one observation from a \emph{Heat Milk} rollout at $\tau=5$\,s, when the window had opened and its reference deadline was about 7.08 simulation seconds away.

\begin{figure*}[t]
\centering
\begin{tcolorbox}[
  enhanced,
  colback=black!2,
  colframe=black!35,
  boxrule=0.5pt,
  arc=3pt,
  left=8pt,
  right=8pt,
  top=6pt,
  bottom=6pt,
  fontupper=\small,
  fontlower=\footnotesize,
  segmentation style={dashed, black!30}
]
\textbf{System Prompt Summary (Native Agent)}\par\smallskip
Control the two PiPER arms using the observations and tools. Submit a target for either or both arms. For each arm, choose six joint angles or an absolute tool pose. Use \texttt{movel} with a tool pose to request straight-line motion. Set joint and gripper speed limits the first time you command each arm. Use the current simulation time and the task window to budget for decisions and motion. If a command fails validation or inverse kinematics, use the diagnostics to revise it from the same observation.
\tcblower
\textbf{Example \texttt{+Info} Observation JSON (\emph{Heat Milk}, $\tau=5$\,s)}\par\smallskip
\begin{verbatim}
{
  "window_information": {
    "as_of_sim_s": 29.90000155940652,
    "opened_windows": [{
      "reference_state": "within",
      "reference_remaining_s": 7.07774793840786
    }]
  }
}
\end{verbatim}
\end{tcolorbox}
\caption{\textbf{Native agent system prompt summary and \texttt{+Info} observation format.} Top: summary of the control instructions provided in the system message. Bottom: excerpt of the \texttt{window\_information} JSON payload attached to an observation in the \texttt{+Info} condition.}
\label{fig:supp:native-prompt}
\end{figure*}
\section{Scene Construction, Physical Processes, and Validation}
\label{sec:scene:details}

\subsection{Scene and Asset Construction Pipeline}
\label{sec:supp:construction-pipeline}

\paragraph{Pipeline overview.}
GPT-6 Astra runs the scene construction pipeline. Given a task specification and either reference photographs or a layout description, the agent plans the layout, calls asset generation tools, builds and refines assets in Blender, and places objects and configures their physical properties for simulation. The pipeline produces a self-contained USD scene package, a task implementation, and a binding configuration. Once constructed, environment loading, physical simulation, and judging execute from these deliverables without further generative model calls. For backgrounds, we prioritize visual realism and plausible spatial arrangements. For objects that the robot interacts with, their function and motion determine the modeling approach so that critical geometry supports the intended physical interactions. The workflow is packaged as reusable agent skills for layout, asset geometry and physics, materials, and task assembly; any agent framework that can load the instructions and call the required tools can reuse them.

\begin{figure*}[t]
\centering
\includegraphics[width=0.86\textwidth]{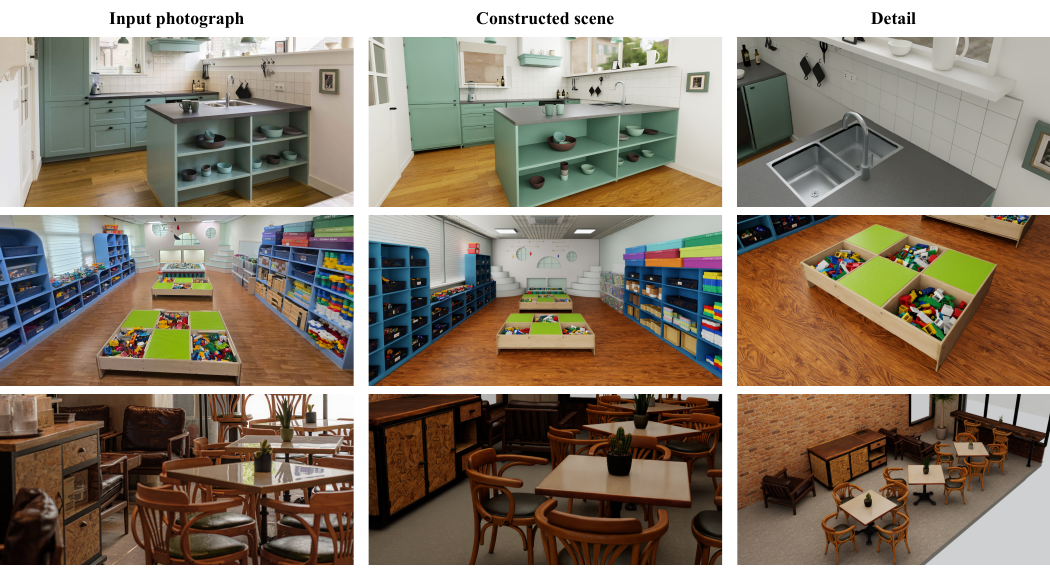}
\caption{\textbf{Generated scenes from reference photographs.} A kitchen, a kindergarten, and a caf\'e. Columns show the input photograph, simulator rendering, and a detail or alternate view.}
\label{fig:scene}
\end{figure*}

\paragraph{Scene layout and background construction.}
We turn each scene reference into a metric layout that specifies the main objects, their counts, and their spatial relationships. We reserve space for the robot mount, task support surfaces, and the manipulation workspace, and record the dimensions, positions, and orientations as explicit 3D parameters. When starting from layout descriptions or supplementing viewpoints, image generation provides scene views to guide layout, lighting, and material choices; it also produces isolated object views for asset modeling. We build the walls, floors, and other basic room geometry in Blender, then place separately modeled furnishings and background objects through USD references. Figure~\ref{fig:scene} shows three resulting backgrounds.

\paragraph{Structure-aware asset modeling.}
We first divide objects into modeling units along their mechanical boundaries. Parts that move relative to one another are modeled separately, with their mounting relationships retained for assembly; fixed surface features stay with their parent component. The geometry that each unit must preserve then determines its modeling route:

\begin{enumerate}
\item \textbf{Whole-asset generation.}
For an independent object without precise mating constraints, TRELLIS.2~\cite{trellis2} generates geometry and physically based rendering (PBR) materials from an isolated reference image. We import the output into Blender to set its scale, origin, and orientation.

\item \textbf{Refinement from a structural model.}
For a body with functional openings, cavities, or mating surfaces, we build a coarse mesh in Blender to set its critical dimensions and structure. UltraShape~\cite{ultrashape} refines the mesh using the reference image. We restore the result to metric scale and remesh it while preserving the required openings, connectivity, and fit. The separate TRELLIS.2 texturing pipeline then adds UV coordinates and PBR materials to the fixed surface.

\item \textbf{Direct modeling and texturing.}
We model hinges, slides, and other parts whose shapes are set by their mating geometry directly in Blender. Where needed, we apply the same texturing pipeline, keeping the modeled shape, position, and scale.
\end{enumerate}

Blender controls dimensions and structural constraints, UltraShape refines geometry conditioned on a reference image, and TRELLIS.2 generates either a complete textured asset or materials for a fixed mesh. Because this component-level pipeline is not restricted to tabletop task objects, it also constructs articulated background furnishings, such as hinged kitchen cabinet doors, with functional joints and physical responses. Although AWAIT's 18 tasks focus on tabletop manipulation and do not interact with these background joints, preserving them allows the generated USD environments to be reused for future mobile or whole-room manipulation tasks.

\begin{figure*}[t]
\centering
\includegraphics[width=\textwidth]{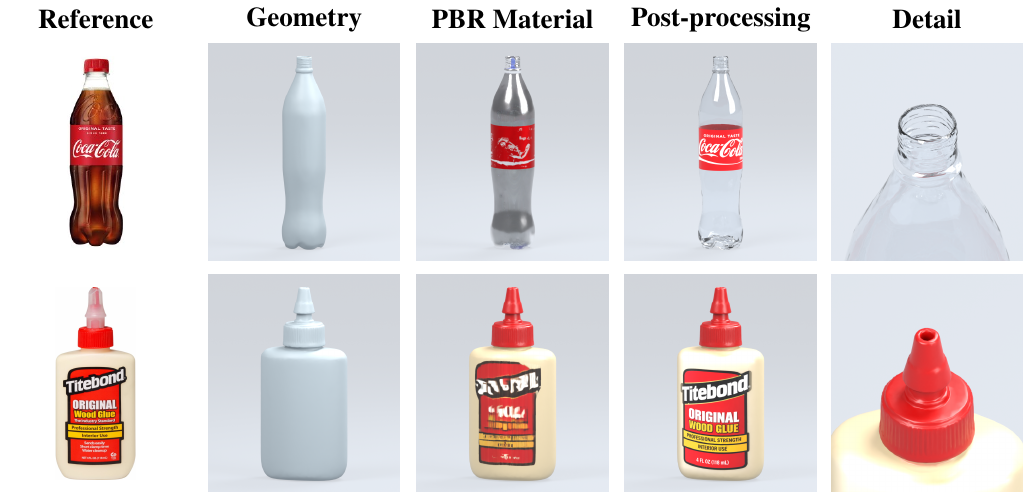}
\caption{\textbf{Appearance post-processing for task assets.} The geometry and PBR columns show the same mesh for each bottle: a structurally modeled cola bottle textured with the fixed-mesh TRELLIS.2 pipeline (top) and a whole-asset TRELLIS.2 generation for the glue bottle with its original generated textures (bottom). Post-processing adjusts USD material attributes to restore the cola bottle's transparent plastic appearance and replaces the illegible generated labels on both bottles with newly generated 2D artwork. The close-ups highlight the functional openings preserved for interaction: the threaded mouth for pouring (top) and the hollow nozzle tip for glue deposition (bottom).}
\label{fig:supp:asset-processing}
\end{figure*}

\paragraph{Appearance refinement.}
We refine the appearance of each asset before simulation. In Blender, we adjust texture scale, orientation, and continuity, as well as surface properties such as roughness and surface normals; broad room surfaces use textures or procedural materials whose channels are baked into USD material bindings. Because raw generative textures can leave small text illegible or bake opaque shading onto translucent parts, post-processing also replaces signs and product labels with newly generated 2D artwork applied via local projections and masks, and directly adjusts USD material attributes (such as transmission and optical properties) to restore transparent materials. Figure~\ref{fig:supp:asset-processing} illustrates this workflow on two task bottles, showing the recovered transparent plastic and legible labels alongside preserved functional openings. For example, to restore the cola bottle's PET shell, the construction agent inspects successive renders and iteratively tunes its USD material using NVIDIA's OmniSurface template in thin-walled mode, with specular transmission enabled, a refractive index of 1.46, and a specular roughness of 0.025, allowing light to pass through the shell while keeping the label opaque.

\paragraph{Physics setup.}
We export assets to USD with their functional openings and cavities intact, assign rigid-body dynamics and colliders to movable bodies, add static colliders to fixed structures, and connect articulated parts with joints. The construction agent estimates mass and contact parameters from object dimensions, material categories, and engineering references such as MatWeb~\cite{matweb}; these values are recorded in USD.

\paragraph{Task assembly and runtime binding.}
We combine the environment and task assets with the standard robot and camera rig, place task objects within the permitted randomization ranges, and fix the layout and task parameters for each instance. Semantic roles, such as the source bottle or receiving cup, are mapped to specific USD prims, binding the task mechanisms (Sec.~\ref{sec:supp:physical-processes}) and judge to the assembled scene prior to the first physics step. Appearance, support, collisions, and physical responses are then verified in the assembled Isaac Sim instance (Sec.~\ref{sec:supp:instance-validation}).

\subsection{Task mechanisms and state changes}
\label{sec:supp:physical-processes}

Task-specific process models couple physical interactions to the visible cues and window predicates used in evaluation. While the host simulates robot contact and object motion, internal process variables such as temperature, film lifetime, and formation progress remain hidden from the evaluated methods. Table~\ref{tab:supp:process-models} gives the state updates and parameters for eight of these mechanisms.

\paragraph{Liquid transfer and absorption.}
\emph{Pour Liquid}, \emph{Clean Mirror}, and \emph{Heat Milk} use particle liquids advanced by the PhysX position-based dynamics solver. In \emph{Pour Liquid}, particle positions relative to the moving bottle and cup determine the volumes in the source, in the receiver, and in transit; liquid that has spilled is counted separately. In \emph{Clean Mirror}, the pump emits particles onto the mirror, and the sponge removes particles within its absorption region at a limited rate. In \emph{Dispense Water}, a reduced flow model advances parcels carrying liquid volume. Parcels entering the cup increase its fill level; missed or excess volume counts as spillage. The model retains liquid in transit after the supply is switched off.

\paragraph{Changes in temperature, shape, and material state.}
In \emph{Heat Milk}, heating updates a tracked liquid temperature that drives thermal expansion and surface foam formation. Once the milk reaches the lower bound of the target temperature range, PhysX diffuse particles near the reconstructed liquid surface seed clusters of translucent bubbles that drift with the local flow and coalesce into a wet foam layer; escaped liquid particles detect overflow. In \emph{Inflate Balloon}, valve state and neck sealing govern gas volume, which sets the rendered size and the target and burst thresholds. In \emph{Flip Hourglass}, a conserved amount of sand transfers between chambers at a rate determined by tilt, updating the rendered sand piles. In \emph{Blow Bubbles}, withdrawing the ring from the solution initiates an aging film whose lifetime governs the active interval. While the film persists, holding the ring within the fan's downstream airflow region at a valid incidence angle accumulates a continuous bubble-formation state that decays otherwise; reaching the detachment threshold releases a bubble that drifts downwind. In \emph{Glue Handle}, deposited patches define the wet working interval, and coverage, alignment, and sustained contact are tracked before a bond is formed.

\begin{table*}[!tb]
\centering
\small
\setlength{\tabcolsep}{4pt}
\renewcommand{\arraystretch}{1.15}
\caption{\textbf{Quantitative process models.} Summary of the state-update equations and physical parameters governing the eight task-specific process models in Sec.~\ref{sec:supp:physical-processes}, where $\Delta t$ denotes one physics step in simulation seconds.}
\label{tab:supp:process-models}
\begin{tabular*}{\textwidth}{@{\extracolsep{\fill}}p{0.15\textwidth}p{0.44\textwidth}p{0.35\textwidth}@{}}
\toprule
Task & State and update & Key parameters \\
\midrule
Pour Liquid & Classify particles relative to the moving bottle and cup; a region containing $N$ particles has volume $N v_p$. & $250\,\mathrm{mL}$ initial liquid across $14{,}224$ particles ($v_p = 250/14224 \approx 0.0176\,\mathrm{mL}$). \\
\addlinespace
Clean Mirror & Emit particles from the spray nozzles when triggered; remove particles inside the sponge's moving absorption box up to the per-step absorption limit. & Mean spray rate $12.5\,\mathrm{mL/s}$; total dose $30\,\mathrm{mL}$; absorption limit $30\,\mathrm{mL/s}$; absorption box extends $2.5\,\mathrm{mm}$ beyond sponge bounds. \\
\addlinespace
Dispense Water & Flow $q \gets q_c + (q - q_c)e^{-\Delta t/t_r}$. Each step emits a parcel of volume $q\Delta t$ falling under gravity; parcels entering the cup add to its fill volume. & Commanded rate $q_c = 15\,\mathrm{mL/s}$ when on ($0$ when off); response time constant $t_r = 0.12\,\mathrm{s}$. \\
\addlinespace
Heat Milk & $T \gets T + r_h\Delta t$ when powered on inside the heating zone (constant otherwise). Particle injection expands volume toward $V_{\mathrm{target}}(T) = 420 + 1180[\max(0,T-65)/35]^2\,\mathrm{mL}$ at rate $\le 180\,\mathrm{mL/s}$; foam seeds at $60\,^\circ\mathrm{C}$ and overflow triggers once escaped volume reaches $1\,\mathrm{mL}$. & Initial temperature $T_0 = 40\,^\circ\mathrm{C}$; heating rate $r_h = 2.4\,^\circ\mathrm{C/s}$ (cooling rate $0\,^\circ\mathrm{C/s}$); pan capacity $1370.4\,\mathrm{mL}$. \\
\addlinespace
Inflate Balloon & Seated with valve pressed: $V \gets V + Q\Delta t$. Unsealed off seat: $V \gets \max(V_0, V e^{-\lambda\Delta t})$. Equivalent radius $r = (3V/4\pi)^{1/3}$; balloon bursts at $r_{\mathrm{burst}}$. & $Q = 650\,\mathrm{mL/s}$; $\lambda = 1.5\,\mathrm{s}^{-1}$; initial radius $r_0 = 2.6\,\mathrm{cm}$ ($V_0 \approx 73.6\,\mathrm{mL}$); target radius $9.5\,\mathrm{cm}$; burst radius $r_{\mathrm{burst}} = 14.5\,\mathrm{cm}$. \\
\addlinespace
Flip Hourglass & Transfer sand from the upper to lower chamber at fractional rate $\bigl[\max(0,(|\cos\theta|-0.15)/0.85)\bigr]^{1.4}/T_d$, capped by available sand, where $\theta$ is the tilt angle from vertical. & Upright drain time $T_d = 14\,\mathrm{s}$; initial sand split $0\%$ upper / $100\%$ lower. \\
\addlinespace
Blow Bubbles & While the film is active, formation progress $p \in [0,1]$ updates as $p \gets p + \Delta t/1.35$ in valid fan airflow and $p \gets p - 2\Delta t$ otherwise; a bubble detaches at $p = 1$. & Film lifetime $20\,\mathrm{s}$ after clearing the solution; valid airflow zone $3.5\text{--}26\,\mathrm{cm}$ downstream, $\le 11\,\mathrm{cm}$ from fan axis, $\ge 65^\circ$ ring-to-flow angle. \\
\addlinespace
Glue Handle & Project the nozzle tip onto the board to deposit glue patches; count patches under the handle's contact footprint and advance the bond timer during continuous valid contact. & Patch radius $2.8\,\mathrm{mm}$; minimum spacing $3.5\,\mathrm{mm}$ between patches; wet working lifetime $30\,\mathrm{s}$ from first in-target deposit. \\
\bottomrule
\end{tabular*}
\end{table*}

\paragraph{Screen updates.}
In \emph{Enter Code}, the display uses pre-authored USD glyphs for each digit slot and status message. Pressing the request button reveals the instance's code and starts the countdown, while the task script updates glyph visibility and status messages as inputs and remaining time change. In \emph{Piano Tiles}, the first valid screen contact starts playback. At each physics step, the script computes the visible tile rectangles from the chart and elapsed time, draws them into a dynamic screen texture, and recolors struck notes green and struck blank tiles red. The same rectangles map contacts on the continuous screen collider to the corresponding beat and lane.

\paragraph{Spring-driven targets.}
In \emph{Whack-a-mole}, each mole is a dynamic rigid body constrained to a vertical prismatic joint with a damped spring drive. For each scheduled appearance, the script ramps the drive target slightly beyond the end of the stroke so that the spring holds the mole extended while still compressing under downward contact. Once fully extended, the mole remains raised for its exposure interval before the drive target returns to zero and retracts it, or retracts immediately upon a valid press.

\subsection{Instance validation}
\label{sec:supp:instance-validation}

\paragraph{Manual review and instance acceptance.}
In the assembled scene, we inspect layout and appearance, paying particular attention to support relationships, unintended intersections, and occlusion of the task workspace. We review materials at the task viewing distance and in close-up to check texture alignment, seams, and transparency. For objects and equipment that participate directly in task interactions, the authors also check four aspects that affect manipulation feasibility and completion time:

\begin{enumerate}
\item \textbf{Physical parameter plausibility.}
We review mass, friction coefficients, and other physical parameters against the object's dimensions, material category, and available engineering references. The values and their supporting assumptions are checked for consistency with the object's structure and expected contact behavior.

\item \textbf{Collision geometry.}
After importing the asset into the scene, we inspect the spatial correspondence between its collision volume and visible geometry. Particular attention is paid to contact surfaces, functional openings, and cavities, so that the collision representation supports the intended interaction.

\item \textbf{Interactive tests of movable components.}
For assets with movable or articulated parts, we apply external forces through Isaac Sim's interactive controls to press, rotate, or open and close the components. We inspect motion direction, travel, joint constraints, and the stability of the connections. Examples include depressing an inflation valve and pressing a button or retractable mole mechanism. These tests assess whether the response to external forces is consistent with the component's function.

\item \textbf{Robot execution.}
For each task instance, an agent uses privileged information, including object poses read from the simulator, to construct a robot action sequence. The sequence is executed with physics and task mechanisms enabled. The authors inspect contact responses, object motion, and component connections under forces applied by the robot, checking that individual interactions remain plausible throughout the complete manipulation sequence. To pass, the execution must meet the task's goal and timing conditions and earn the full episode score before the episode horizon.
\end{enumerate}

Issues identified during review are addressed by revising the relevant geometry, collision representation, physics parameters, or installation pose. The affected checks are then repeated. Each instance must pass the complete task execution check before acceptance.

\FloatBarrier
\section{Complete Task Specifications and Delay Grids}
\label{sec:supp:tasks}

\subsection{Controlled-delay grid}
\label{sec:exp:delay-grid}

Table~\ref{tab:supp:delay-grid} lists the five controlled delays $\tau_1,\dots,\tau_5$ and the reference-window length $W^\star$ for each task; all method--VLM configurations share the same grid. To compute $W^\star$, we extract event-opening states from each task's fixed, accepted privileged validation trajectories (Sec.~\ref{sec:supp:instance-validation}) and measure each event's duration along the prescribed no-intervention reference continuation. Event durations are averaged within each instance, and instances are then weighted equally to obtain a single task-level $W^\star$ shared across methods, models, and timing conditions, independently of evaluated rollout outcomes.

In \emph{Catch Ruler}, the gripper starts in position to catch the ruler by closing its fingers alone. To test reaction time in this setup, we deliberately choose a maximum delay ($\tau_5=0.256$\,s) close to the reference-window length $W^\star$.

\begin{table}[!tb]
\centering
\small
\setlength{\tabcolsep}{3pt}
\caption{Controlled delays per high-level decision round and reference-window lengths by task, in simulation seconds. Reference-window lengths $W^\star$ are rounded to two decimal places.}
\label{tab:supp:delay-grid}
\begin{tabular}{lrrrrrr}
\toprule
Task & $\tau_1$ (s) & $\tau_2$ (s) & $\tau_3$ (s) & $\tau_4$ (s) & $\tau_5$ (s) & $W^\star$ (s) \\
\midrule
Withdraw Cash    & 0.35  & 0.7   & 1.4   & 2.8   & 5.6 & 11.00 \\
Glue Handle      & 0.5   & 1     & 1.5   & 5     & 10 & 30.00 \\
Enter Code       & 0.5   & 1     & 1.5   & 3     & 10 & 30.00 \\
Clean Mirror     & 0.5   & 1     & 2     & 6.5   & 10 & 25.00 \\
Stamp Paper      & 0.5   & 1     & 2     & 3     & 5 & 15.00 \\
Blow Bubbles     & 0.5   & 1     & 2     & 5     & 10 & 20.00 \\
\midrule
Inflate Balloon  & 0.5   & 1     & 2     & 4     & 8 & 14.12 \\
Dispense Water   & 0.4   & 0.8   & 2     & 6     & 12 & 16.00 \\
Flip Hourglass   & 0.45  & 0.9   & 1.8   & 3.6   & 5 & 13.69 \\
Conveyor Picking & 0.2   & 0.4   & 2     & 4     & 16 & 36.72 \\
Heat Milk        & 0.5   & 1     & 2     & 3     & 5 & 15.18 \\
Pour Liquid      & 0.2   & 0.4   & 1     & 2     & 6 & 12.90 \\
\midrule
Stop Dominoes    & 0.075 & 0.1   & 0.2   & 0.4   & 0.8 & 1.39 \\
Whack-a-mole     & 0.075 & 0.15  & 0.3   & 0.6   & 1.2 & 2.35 \\
Catch Ruler      & 0.008 & 0.016 & 0.032 & 0.064 & 0.256 & 0.25 \\
Piano Tiles      & 0.3   & 0.6   & 1.2   & 2.4   & 4 & 7.50 \\
Catch Cup        & 0.02  & 0.04  & 0.08  & 0.2   & 0.4 & 0.78 \\
Goalkeeping      & 0.02  & 0.04  & 0.08  & 0.2   & 0.4 & 1.05 \\
\bottomrule
\end{tabular}
\end{table}

\subsection{Task specifications and instance randomization}
\label{sec:scene:task-specifications}

\paragraph{Instance randomization.}
Table~\ref{tab:supp:task-catalog} specifies each task's prompt, action window, and full-score criteria. By default, all tasks randomize object poses across instances subject to the spatial constraints below. Additional task-specific variations that do not alter the target action-window duration, such as onset times, target content, or event orders, are listed under each task's \emph{Instance variation} entry in Table~\ref{tab:supp:task-catalog} (omitted for tasks that vary only object poses). Although the construction workflow supports modifying process speeds, lifetimes, and other timing parameters when creating new task variants, the reported evaluation holds all parameters governing window duration fixed. Each instance's layout, background, and task parameters are fixed during construction and stored in its initial scene and task implementation; loading or resetting the instance does not resample them, and paired rollouts share the same configuration.

\paragraph{Spatial constraints.}
Object poses vary within the robot's reachable workspace and the head camera's field of view while preserving the relative geometry that determines the task's temporal dynamics. Objects with such dependencies are assembled under a common USD root and repositioned together. In \emph{Goalkeeping}, for example, the ball and goal share an assembly whose placement can change without altering their initial relative positions. With the prescribed launch velocity fixed in the assembly's coordinates, this preserves the ball's travel distance and reference arrival time at the goal, while the ball remains an independent dynamic body during execution.

\subsection{Event windows and scoring rules}
\label{sec:supp:event-rules}

Table~\ref{tab:supp:task-catalog} specifies the opening and closing conditions for each task. Each reference window is the half-open interval $[t_{\mathrm{open}},t_{\mathrm{close}})$, where $t_{\mathrm{close}}$ is evaluated along the prescribed continuation from the actual opening state without further system intervention, so subsequent robot actions do not alter $t_{\mathrm{close}}$. Fixed lifetimes use an elapsed-time state variable; moving objects and evolving quantities use the corresponding physical or process state. Reference calculations and no-intervention calibration rely on privileged host state that is not exposed to evaluated methods.

\paragraph{Event and episode scores.}
Single-event tasks use a binary score: the prescribed goal must be achieved within its window, including any required contact, alignment, and holding; submitting a control target alone is insufficient, and transient satisfaction does not override task-defined retention or failure checks. Two tasks (\emph{Whack-a-mole} and \emph{Piano Tiles}) contain multiple events and define one window per target appearance. For \emph{Whack-a-mole}, the episode score is $H/6$, where $H$ is the number of valid hits among the six scheduled appearances. For \emph{Piano Tiles}, let $J$ be the number of target notes, $H$ the number hit within their windows, and $Q$ the number of wrong presses. The score is $\max(0,(H-Q)/J)$: a missed target adds no hit, and a wrong press incurs a penalty equal to one hit, so a full score requires hitting all target notes with no wrong presses.

\clearpage
\begingroup
\newcommand{\TaskWindowBand}[1]{%
  \noindent
  {\setlength{\fboxsep}{4pt}%
   \colorbox{black!6}{\parbox{\dimexpr\linewidth-2\fboxsep\relax}{\strut\textsc{#1}}}}%
  \par\smallskip
}
\newcommand{\TaskEntry}[7][]{%
  \par\noindent
  \begin{minipage}{\linewidth}
    \setlength{\parskip}{3pt}%
    \if\relax\detokenize{#1}\relax\else\TaskWindowBand{#1}\fi
    \noindent\textbf{#2}\par
    \noindent\emph{Task prompt.} #3\par
    \noindent\emph{Window opens.} #6\par
    \noindent\emph{Window closes.} #7\par
    \noindent\emph{Full-score criteria.} #4\par
    \if\relax\detokenize{#5}\relax\else
      \noindent\emph{Instance variation.} #5\par
    \fi
  \end{minipage}\par\addvspace{5pt}
}
\tcbset{task frame/.style={enhanced,frame hidden,colback=white,boxrule=0.5pt,arc=0pt,fontupper=\small,
  nobeforeafter,boxsep=0pt,left=9pt,right=9pt,top=9pt,bottom=9pt}}
\newcommand{\TaskFrameTop}{\draw[line width=0.5pt]
  (frame.south west) -- (frame.north west) -- (frame.north east) -- (frame.south east);}
\newcommand{\TaskFrameSides}{\draw[line width=0.5pt]
  (frame.south west) -- (frame.north west) (frame.south east) -- (frame.north east);}
\begin{table*}[p]
\caption{\textbf{Task specifications and action windows.} Each entry gives the task prompt, window boundaries, full-score criteria, and task-specific variation beyond object pose, where applicable.}
\label{tab:supp:task-catalog}
\noindent\begin{tcolorbox}[task frame,overlay={\TaskFrameTop}]
\TaskEntry[Self-triggered windows]{Withdraw Cash}
  {Press the lowest withdrawal button on the left side of the screen. Retrieve the banknote within 11\,s of presentation.}
  {The banknote clears the slot and is held within one gripper, with both fingers of that gripper contacting it.}
  {}
  {The banknote reaches its presented position at the cash slot.}
  {The 11\,s retrieval period after presentation expires.}

\TaskEntry{Glue Handle}
  {Pick up the open glue bottle and apply glue inside the black outline. Then align the handle underside with the outline and press for 2\,s within the 30\,s working period.}
  {Before the glue expires, the aligned handle covers at least six glue deposits in the target region and bonds after 2\,s of continuous pressure above 1.5\,N, measured over a 50\,ms window.}
  {}
  {The first glue deposit lands inside the outline.}
  {The 30\,s working period after the first in-box deposit expires.}

\TaskEntry{Enter Code}
  {Press the request button, then enter the displayed verification code within the time limit.}
  {The entered sequence matches the displayed code before expiry. The device checks the entry automatically when all digits are entered.}
  {The three-digit code, using digits 1--9.}
  {The first accepted request displays the code.}
  {The code expires 30\,s after the request.}

\TaskEntry{Clean Mirror}
  {Grasp the sponge by its top handle and trigger the spray, then wipe the mirror with the sponge underside to absorb the cleaning liquid.}
  {After the full 30\,mL dose has been sprayed, the sponge has absorbed more than 40\% of it and less than 2\,mL remains near the mirror.}
  {}
  {A valid button press triggers spraying.}
  {The 25\,s cleaning period after the trigger expires.}

\TaskEntry{Stamp Paper}
  {Pick up the stamp and ink it. Then stamp inside the marked box within 15\,s.}
  {The stamp face makes physical contact with the paper, with at least 80\% of the face inside the contact band over the paper. The imprint is made while the ink is valid, and at least 50\% of its projected area lies inside the box.}
  {}
  {The stamp first makes valid contact with the ink pad.}
  {The 15\,s ink lifetime elapses.}

\TaskEntry{Blow Bubbles}
  {Pick up the purple bubble wand from the solution and move it in front of the running fan to blow a soap bubble.}
  {A bubble forms and detaches from the ring before the film expires. Formation progress rises when the ring is in the fan stream and properly oriented.}
  {}
  {The immersed ring first clears the liquid surface completely.}
  {The 20\,s film lifetime expires; re-dipping does not reset it.}

\end{tcolorbox}
\end{table*}

\begin{table*}[p]
\ContinuedFloat
\caption{\textbf{Task specifications and action windows (continued).}}
\noindent\begin{tcolorbox}[task frame,overlay={\TaskFrameSides}]
\TaskEntry[Cued windows]{Inflate Balloon}
  {Hold the blue valve until the balloon is wider than 19\,cm and stop the airflow before it bursts. Pinch the soft neck and lift the balloon off while keeping it pinched.}
  {The balloon is detached from the inflator, unburst, wider than 19\,cm, and still pinched at the neck.}
  {}
  {The balloon first exceeds 19\,cm in diameter.}
  {Continued inflation at the prescribed flow reaches the burst volume.}

\TaskEntry{Dispense Water}
  {Touch the lower blue water-drop button to start dispensing. Touch it again when the water reaches the orange arc inside the cup.}
  {The cup is within 120\,mL of the target, less than 50\,mL of dispensed water is outside the cup, and the supply is off with no residual flow.}
  {The target volume; the height of the orange fill mark follows it.}
  {The cup volume first reaches the target minus 120\,mL.}
  {With the cup and nozzle poses held at their values at opening, continued supply is predicted to raise the cup volume above the target plus 120\,mL. The prediction accounts for water already in transit and the supply response.}

\TaskEntry{Flip Hourglass}
  {Use each arm in turn to grasp and invert the hourglass. Finish the second inversion before the sand runs out. During each turn, keep the other arm clear of the hourglass.}
  {Different arms complete the two inversions with exclusive grasps. Each holds the hourglass at the opposite upright orientation, and the second completes before the upper chamber empties.}
  {}
  {The first valid inversion is completed.}
  {The upper chamber is predicted to empty from its sand quantity and drainage rate at opening.}

\TaskEntry{Conveyor Picking}
  {Put the small red block into the tray.}
  {The red block lies in the tray acceptance region according to its footprint and base height. No other block satisfies that criterion.}
  {The red block's position in the release sequence.}
  {The red block is released onto the belt.}
  {The red block is predicted to reach the belt end from its scheduled release time, spawn position, and belt speed.}

\TaskEntry{Heat Milk}
  {Turn on the cooktop and move your hand aside to observe the milk. When new bubbles persist or foam lifts the surface, turn off the cooktop or lift the pot off.}
  {The milk is within the target temperature range (60--98\,$^\circ$C), has been unheated for at least 1\,s, and has never overflowed.}
  {}
  {The milk first reaches the lower limit of the target temperature range.}
  {Under continued heating, the milk first reaches the upper temperature limit or overflows, whichever occurs earlier.}

\TaskEntry{Pour Liquid}
  {Pour until the cup reaches its green fill mark, then return the bottle upright to stop pouring.}
  {The cup is within 80\,mL of the target and varies by at most 2\,mL over 1\,s. Over the most recent 0.5\,s, the source volume changes by at most 1\,mL and liquid in transit stays at or below 1\,mL.}
  {The target volume; the green fill mark follows it.}
  {The cup volume first reaches the target minus 80\,mL.}
  {At the prescribed 10\,mL/s reference pour rate, the cup reaches the target plus 80\,mL or the bottle empties, whichever occurs first given the volumes at window opening.}

\end{tcolorbox}
\end{table*}

\begin{table*}[p]
\ContinuedFloat
\caption{\textbf{Task specifications and action windows (continued).}}
\noindent\begin{tcolorbox}[task frame,overlay={\draw[line width=0.5pt] (frame.north west) -- (frame.south west) -- (frame.south east) -- (frame.north east);}]
\TaskEntry[Sudden-onset windows]{Stop Dominoes}
  {Steady the red domino as soon as the dominoes start falling. Touching the red domino before the dominoes start falling results in a score of zero.}
  {A domino counts as fallen if it is tilted more than $5^\circ$ from upright. After propagation begins, success requires all upstream dominoes and no downstream dominoes to be down. The robot must not touch the red domino before the first domino falls.}
  {The disturbance onset delay (0.2--2\,s).}
  {The mechanism gives the first domino its initial linear and angular velocities.}
  {The red domino topples under the prescribed disturbance and fixed propagation intervals.}

\TaskEntry{Whack-a-mole}
  {Keep watching and strike each mole as soon as it emerges.}
  {Every scheduled appearance is hit by depressing the mole below 50\% of its stroke during its exposure window.}
  {The sequence of hole indices for the six appearances; a hole can appear more than once.}
  {After the rise phase, the mole is first measured at 95\% of its stroke or higher.}
  {The 2.35\,s interval after measured extension expires, even if the mole was hit and retracted earlier.}

\TaskEntry{Catch Ruler}
  {Wait for the ruler to start falling, then quickly grasp it. Touching it before it falls results in a score of zero.}
  {Before release, a finger contact force above 0.15\,N counts as a premature grasp and yields zero. After release, the gripper must contact the ruler and reduce its vertical speed below 2.5\,m/s.}
  {The release onset delay (0.3--1.5\,s).}
  {The latch releases the ruler.}
  {Free fall without intervention first reaches the vertical-speed limit.}

\TaskEntry{Piano Tiles}
  {Tap the screen to start, then press the moving dark piano tiles and avoid the blank tiles.}
  {Every note is hit while visible and no blank tile is hit. A hit requires more than 3\,N of total normal force on that tile.}
  {The lane assignments and blank beats across the fixed eight-beat chart (including whether consecutive notes share a lane), while holding scroll speed fixed.}
  {A target tile first becomes visible after playback starts.}
  {At the prescribed scroll speed, the entire target tile leaves the display.}

\TaskEntry{Catch Cup}
  {The cup is rolling toward the edge of the table. Stop it from falling off.}
  {The robot has contacted the cup, its current translational speed is below 0.05\,m/s, and its center has never dropped by at least 5\,cm from its initial height.}
  {}
  {The episode starts with the cup rolling toward the edge.}
  {Without robot intervention, the cup center first drops by at least 5\,cm from its initial height.}

\TaskEntry{Goalkeeping}
  {Don't move until the ball is kicked. Then use the robot arm to block the ball and keep it out of the goal.}
  {A save requires the robot to have touched the ball after the kick and the ball to be outside the goal with a translational speed below 0.05\,m/s.}
  {The kick onset delay (0--1.0\,s).}
  {The ball is kicked toward the goal.}
  {Without intervention, the entire ball crosses the goal plane through the opening under the rolling model.}

\end{tcolorbox}
\end{table*}
\endgroup

\end{document}